\documentclass{article}

\usepackage{microtype}
\usepackage{graphicx}
\usepackage{subcaption}
\usepackage{booktabs} 

\usepackage{hyperref}

\usepackage[preprint]{icml2026}

\usepackage{amsmath}
\usepackage{amssymb}
\usepackage{mathtools}
\usepackage{amsthm}

\usepackage[capitalize,noabbrev]{cleveref}

\theoremstyle{plain}

\theoremstyle{definition}

\theoremstyle{remark}

\usepackage[textsize=tiny]{todonotes}

\usepackage{graphicx}
\usepackage{multirow}
\usepackage{makecell}
\usepackage{subcaption}
\usepackage{amsmath}
\usepackage{amssymb}

\newcommand{\gu}[1]{{\color{black}#1}}
\newcommand{\yj}[1]{{\color{black}#1}}  

\icmltitlerunning{Continual-MEGA: A Large-scale Benchmark for Generalizable Continual Anomaly Detection}

\begin{document}

\twocolumn[
  \icmltitle{Continual-MEGA: A Large-scale Benchmark for Generalizable Continual Anomaly Detection}



  \icmlsetsymbol{equal}{*}

  \begin{icmlauthorlist}
    \icmlauthor{Geonu Lee}{snuailab}
    \icmlauthor{Yujeong Oh}{snuailab}
    \icmlauthor{Geonhui Jang}{chungang}
    \icmlauthor{Soyoung Lee}{chungang}
    \icmlauthor{Jeonghyo Song}{chungang}
    \icmlauthor{Sungmin Cha}{nyu}
    \icmlauthor{YoungJoon Yoo}{snuailab,chungang}

  \end{icmlauthorlist}

  \icmlaffiliation{snuailab}{SNUAILAB}
  \icmlaffiliation{chungang}{Chung-Ang University}
  \icmlaffiliation{nyu}{New York University}

  \icmlcorrespondingauthor{YoungJoon Yoo}{yjyoo3312@cau.ac.kr}

  \icmlkeywords{Machine Learning, ICML}

  \vskip 0.3in
]



\printAffiliationsAndNotice{}  


\begin{abstract}

We introduce a new benchmark for continual learning in anomaly detection, \gu{addressing the limited adaptability of static models in dynamic real-world deployment scenarios.}
Our benchmark, Continual-MEGA, \gu{expands existing evaluation settings by integrating diverse public datasets} with our newly proposed dataset, ContinualAD.
Beyond standard continual learning settings, we additionally propose a scenario that evaluates zero-shot generalization to unseen classes not encountered during continual adaptation.
This reflects recent advances in continual zero-shot research and highlights its practical significance. 
This setting introduces a new agenda for the anomaly detection field, and we conduct extensive evaluations of various existing anomaly detection algorithms designed for continual or zero-shot scenarios, as well as our proposed baseline methods.
From our experiments, we derive three key findings: (1) existing methods exhibit significant limitations, particularly in pixel-level defect localization, (2) the proposed ContinualAD dataset is effective for the proposed benchmarking scenario, and (3) our baseline method suggests a promising direction for designing CLIP-based continual and generalizable frameworks through simple adaptation combined with feature synthesis.
\end{abstract}

\section{Introduction}
\label{sec:intro}

Anomaly detection (AD)~\citep{roth2022towards, liu2023simplenet, jeong2023winclip, zhou2023anomalyclip} plays a crucial role in quality control, ensuring precise identification of defects during production. 
It is widely applied in automated detection across diverse domains, including industrial and agricultural products, medical images, and other application areas~\citep{huang2024adapting, wei2025deep}. 
Due to the complexity and variety of real-world environments, anomaly detection models need to recognize a wide range of defects~\citep{guo2025two}.
To tackle this issue, several benchmark scenarios have been established using public datasets~\citep{bergmann2019mvtec,zou2022spot,mishra2021vt,wang2024real,jezek2021deep,lehr2024ad3}. These datasets, such as the widely used MVTec-AD~\citep{bergmann2019mvtec} and VisA~\citep{zou2022spot}, encompass a wide range of object categories, including fabrics, food items, consumer goods, and industrial components.

Conventional deep approaches~\citep{bergmann2022beyond,cohen2020sub,defard2021padim,li2021cutpaste,ristea2022self,roth2022towards,zavrtanik2021draem,zou2022spot} assume unsupervised or per-class anomaly detection.
Following the advancement of CLIP~\citep{radford2021learning} and its initial application to AD~\citep{jeong2023winclip}, unified AD frameworks~\citep{you2022unified,he2024learning,yao2024hierarchical,strater2024generalad} have emerged, allowing a single model to handle various evaluation scenarios. These approaches include continual learning and adaptation~\citep{li2022towards, pang2025context, liu2024unsupervised, tang2024incremental, jin2024oner,meng2024moead,mcintosh2024unsupervised} as well as zero-shot, few-shot AD~\citep{jeong2023winclip, zhou2023anomalyclip, li2024promptad, zhu2024toward, deng2023anovl, chen2024clip, chen2023zero, tamura2023random, gu2024filo,gui2024few,qu2024vcp}.

From a dataset perspective, \gu{the difficulty of collecting defect samples} makes AD more challenging than general vision tasks~\citep{fang2023fastrecon}.
Consequently, widely used evaluation datasets~\citep{bergmann2019mvtec,zou2022spot} are significantly limited in both size and variability. 
This limitation has motivated recent research to explore continual~\citep{liu2024unsupervised}, zero-shot~\citep{zhou2023anomalyclip}, and few-shot learning~\citep{zhang2024mediclip} settings as strategies to overcome data scarcity.
In light of this limitation, we suggest the need for new evaluation benchmarks with a larger quantity of data, achieved by integrating diverse public datasets and curating additional samples to increase both the volume and the variety of data.

\begin{figure}[t]
\centering
\includegraphics[width=0.95\linewidth]{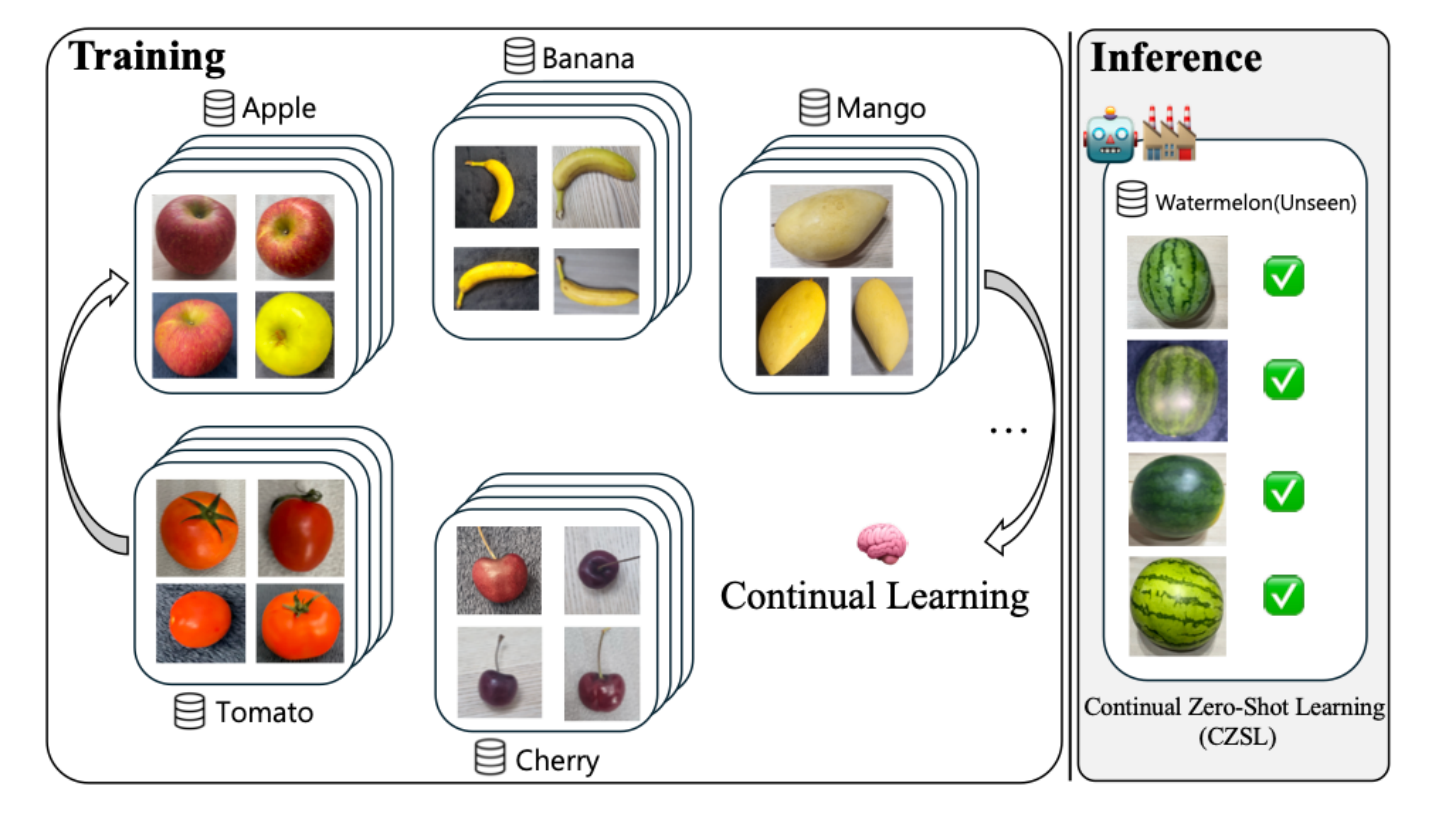} 
\vspace{-0.5em}
\caption{\textbf{Motivation for using Continual Learning (CL) and Continual Zero-Shot Learning (CZSL)} in anomaly detection, enabling models to handle evolving defects over time and generalize to unseen anomalies without retraining.}
\vspace{-2em}
\label{fig:cl_czsl}
\end{figure}

Due to these limitations, anomaly detection systems deployed in real-world environments often face sequentially arriving tasks, 
where new object categories or defect types emerge over time~\citep{li2022towards, liu2024unsupervised}. 
In such scenarios, retraining models from scratch for every new task is \gu{not only} computationally expensive \gu{but often infeasible due to strict data retention policies or privacy regulations that prohibit storing long-term historical data}~\citep{wang2024comprehensive, liu2024unsupervised}. 
\gu{Furthermore, the extreme rarity of anomaly samples makes it difficult to re-collect identical defect patterns from the past, rendering catastrophic forgetting a critical failure in maintaining factory-wide safety~\citep{bugarin2024unveiling}.}
Continual learning aims to address these challenges by enabling the models to incrementally adapt to new data while preserving knowledge of previously seen tasks. 
However, in many practical cases, some tasks or defect types may not be observed during training at all, 
which requires models to generalize to the entirely unseen classes (i.e., tasks or defect types)~\citep{zhou2023anomalyclip}. 
This setting, known as continual zero-shot learning (CZSL)~\citep{zhang2023continual}, 
is particularly crucial to building robust and scalable anomaly detection systems, as illustrated in Figure~\ref{fig:cl_czsl} with a real-world example.

In this paper, we introduce a novel and comprehensive evaluation benchmark, \textbf{Continual-MEGA}, that is designed to evaluate the continual and zero-shot capabilities of anomaly detection models. Our benchmark includes a large-scale evaluation dataset that integrates widely used public datasets~\citep{bergmann2019mvtec,zou2022spot,mishra2021vt,wang2024real,jezek2021deep,lehr2024ad3} with a newly curated dataset, \textbf{ContinualAD}.
The Continual-MEGA benchmark supports two primary evaluation scenarios: (1) a standard continual learning setup, and (2) an extended setup evaluating generalization performance after the continual learning phase, often required in real-world applications. 
This second setting aligns with the concept of continual zero-shot learning (CZSL)~\citep{zhang2023continual}, where models are expected to adapt to new tasks incrementally while maintaining strong generalization to entirely unseen classes.
We highlight that the development of zero-shot unified AD models, an increasingly popular direction, naturally includes addressing a stream of continually incoming novel objects, which corresponds to the CZSL application scenario, as illustrated in Figure~\ref{fig:cl_czsl}.

We conduct extensive evaluation on the proposed Continual-MEGA benchmark, testing representative anomaly detection methods~\citep{cao2024adaclip,huang2024adapting,liu2024unsupervised,liu2023simplenet,qu2024vcp, strater2024generalad,tang2024incremental,tao2024kernel,yao2024hierarchical,zhang2024mediclip,zhou2023anomalyclip} and clearly demonstrating that substantial room for improvement remains for the AD domain in terms of continual adaptation and generalizability.
Furthermore, through the proposed baseline method, Anomaly Detection across Continual Tasks (ADCT), aligning with the evaluation results, we show that the method employing minimal adapters with feature synthesis applied to the pretrained CLIP backbone achieves overall superior performance, suggesting that excessive
adaptation and guidance to CLIP may lead to overfitting on previously seen objects.
Our contributions are summarized as follows:

\begin{itemize}    
    \item
    We introduce \textbf{Continual-MEGA}, a novel large-scale continual learning benchmark, featuring detailed evaluation scenarios. The benchmark is constructed by integrating existing public datasets with our newly curated dataset, \textbf{ContinualAD}, which notably expands the overall data volume and diversity.
    \item 
    From extensive evaluations on the proposed Continual-MEGA benchmark, we demonstrate that there is substantial enough room for improvement in AD performance.
    \item 
    We introduce a baseline method, \textbf{Anomaly Detection across Continual Tasks (ADCT)}, which integrates lightweight MoE-style adapter modules and anomaly feature synthesis with CLIP, pointing toward a promising direction for both Continual and CZSL scenarios.
\end{itemize}

\section{Related Works}
\label{sec:related_works}

Anomaly detection focuses on detecting and rejecting unknown samples~\citep{amodei2016concrete, hendrycks2021unsolved}, framed as an out-of-distribution (OOD) problem. 
Recent advances in large-scale backbone models, such as CLIP~\citep{radford2021learning}, offer a promising solution to the challenge of unified AD across categories~\citep{jeong2023winclip,he2024learning,yao2024hierarchical,strater2024generalad}. 
Following the initial approach~\citep{jeong2023winclip} using CLIP, current research trends focus on developing unified anomaly detection models with zero- and few-shot category adaptation.

\paragraph{Zero- and Few-shot Adaptation.}
{Anomaly detection with zero- and few-shot adaptation~\citep{jeong2023winclip,li2024promptad, chen2023zero, tamura2023random, gu2024filo,gui2024few,qu2024vcp,zhou2023anomalyclip,deng2023anovl,chen2024clip} across various categories reflects real-world scenarios where acquiring a sufficient number of samples for newly incoming categories is often infeasible}, and obtaining anomaly samples is even more challenging. 
Various methods have been proposed to address these challenges, including text prompt utilization~\citep{jeong2023winclip, li2024promptad, zhou2023anomalyclip, deng2023anovl, chen2024clip, tamura2023random, gu2024filo}, visual context prompting~\citep{qu2024vcp, deng2023anovl}, and anomaly dataset synthesis~\citep{chen2023zero,chen2024unified}. Notably, most of these approaches leverage text prompt information, with strategies ranging from manually designed templates~\citep{jeong2023winclip, deng2023anovl}, normal-sample-only strategies~\citep{li2024promptad}, learned prompts~\citep{zhou2023anomalyclip, deng2023anovl, gu2024filo}, to augmented prompting techniques~\citep{tamura2023random}.

\paragraph{Continual Adaptation.}
Another notable trend in recent anomaly detection research is continual adaptation~\citep{li2022towards, pang2025context, liu2024unsupervised, tang2024incremental, jin2024oner,meng2024moead,mcintosh2024unsupervised}, where object categories arrive incrementally. In this context, the primary goal is to mitigate catastrophic forgetting while ensuring that adaptation to previous categories improves the model's performance for future category adaptations.
Based on initial efforts~\citep{li2022towards}, many approaches have been proposed, including context-aware feature adaptation~\citep{pang2025context}, learned text prompts~\citep{liu2024unsupervised}, unified reconstruction-based detection frameworks~\citep{tang2024incremental}, online replay memory mechanisms~\citep{jin2024oner}, parameter-efficient tuning strategies~\citep{meng2024moead}, and unsupervised tuning approaches~\citep{mcintosh2024unsupervised, tang2024incremental}. The continual evaluation scenario is built on public datasets such as MVTec-AD~\citep{bergmann2019mvtec} or VisA~\citep{zou2022spot}, but the quantity and diversity of the dataset are limited compared to the scenario~\citep{bang2021rainbow} using ImageNet~\citep{deng2009imagenet}. 

\paragraph{Continual Zero-shot Learning.}
Continual zero-shot learning (CZSL)~\citep{zhang2023continual,chaudhry2018efficient} has recently emerged as a paradigm that aims to simultaneously preserve past knowledge and \yj{adapt} to future tasks, mirroring the way humans learn throughout their lifetime. Subsequent studies have addressed these problems by generative replay~\citep{gautam2024generative}, context composition~\citep{zhang2024continual}, and class normalization~\citep{skorokhodov2020class}.
Recognizing the practical relevance of the CZSL setting in AD applications, we construct a large-scale Continual-MEGA benchmark designed to incorporate CZSL scenarios.


\begin{figure*}[t]
\centering
\includegraphics[width=0.8\textwidth]{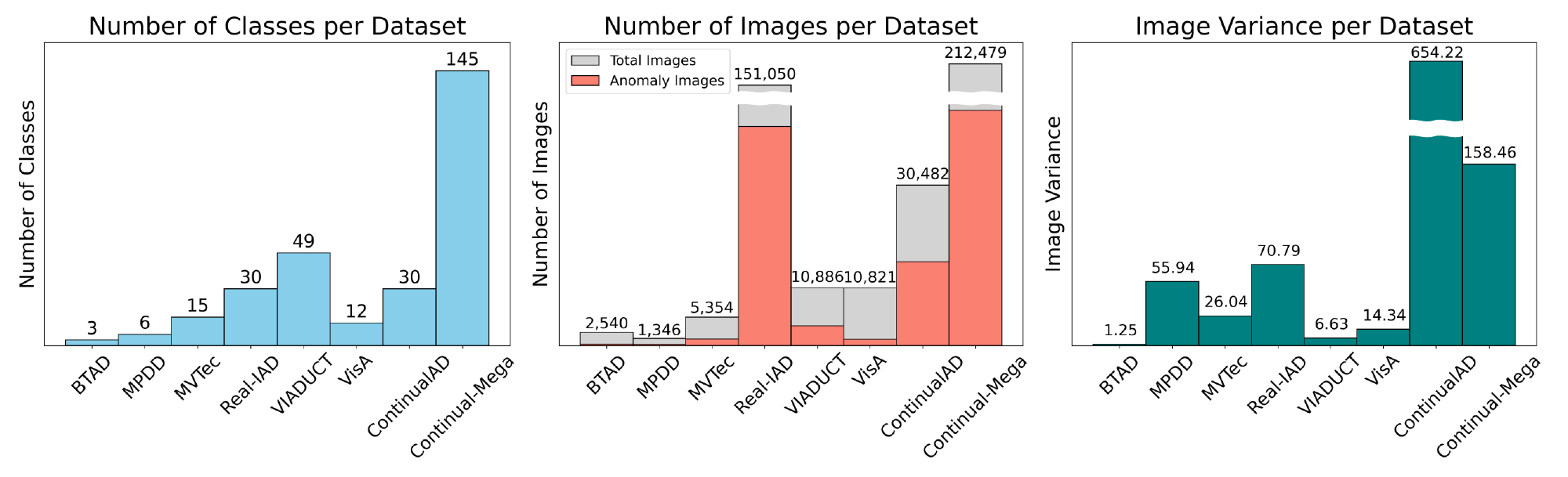} 
\vspace{-1.0em}
\caption{\textbf{Illustration of statistics of various datasets.} Each graph (from left to right) shows the number of classes, number of images, and pixel value variance for public datasets, as well as our proposed ContinualAD and Continual-MEGA. Image variance is defined as the \textbf{average per-pixel variance} for each class.}
\vspace{-1.2em}
\label{fig:dataset_statistics}
\end{figure*}

\begin{figure*}[t]
\centering
\includegraphics[width=0.7\textwidth]{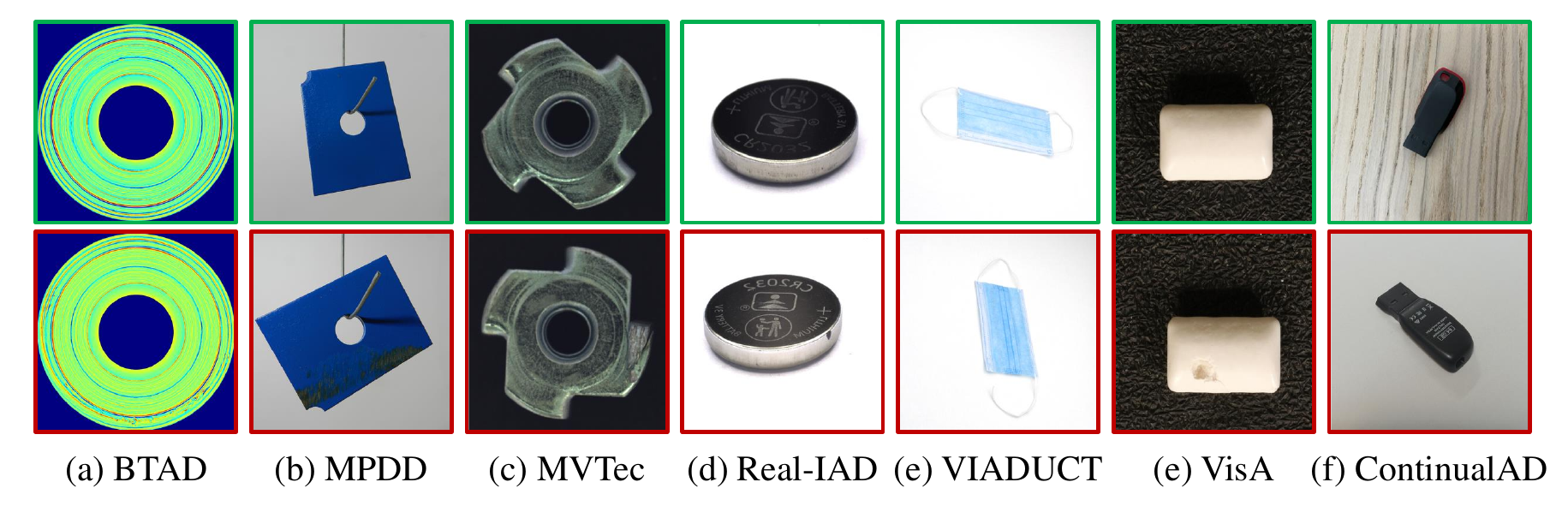} 
\vspace{-0.5em}
\caption{\textbf{Example visualizations of sample images} from various public anomaly detection datasets and the proposed ContinualAD dataset. Green boxes indicate normal images, while red boxes represent anomaly images.}
\vspace{-1.5em}
\label{fig:example_images}
\end{figure*}

\section{Continual-Mega Benchmark}
\subsection{ContinualAD Dataset} To form the Continual-MEGA benchmark, we propose the ContinualAD dataset, which is \yj{significantly larger in scale compared to previous datasets.} 
The ContinualAD dataset comprises 30 classes; a detailed breakdown of the normal and anomaly sample counts for each category is summarized in Table~\ref{tab:cls_counts} in the Appendix.
Moreover, the ContinualAD dataset includes a wide range of object instances within the same class, enabling the evaluation of robustness to intra-class variation.
Consequently, as illustrated in Figure~\ref{fig:dataset_statistics}, the ContinualAD dataset exhibits significantly higher image variance than existing datasets. The variance value is computed by first calculating the pixel-wise variance across images within each class and then averaging these values across all classes in the dataset. This indicates that prior datasets primarily contain visually similar images within each class, limiting their ability to evaluate model performance under diverse conditions. In contrast, the higher variance in ContinualAD facilitates more realistic and challenging evaluation scenarios.

\paragraph{Dataset Acquisition.}
The proposed ContinualAD dataset is constructed from 30 real-world object categories, for which normal images were first collected under normal conditions. 
For each object, we then deliberately induced diverse defects such as cracks, holes, rot, scratches, bending, and contamination to obtain corresponding anomaly images. 
All images were captured using 10 devices (Galaxy S21+, iPhone 12 Pro Max, iPhone 13, iPhone 15 Pro Max, iPhone XS, iPad Air 4, iPad Pro 11-inch 2nd generation, iPad Pro 12.9-inch 4th generation, iPhone 12 mini, and ZFLIP 3), and the anomaly regions in the anomaly images were annotated at the pixel level using polygon masks. 
This diversity in devices and capturing conditions enables evaluation under a wide range of real-world anomaly scenarios and environmental variations.

\paragraph{Dataset Statistics.}
Figure~\ref{fig:dataset_statistics} further summarizes the overall class distribution and the number of samples per class. The graph highlights that the ContinualAD dataset offers a competitively large number of classes and samples compared to existing Real-IAD dataset. Notably, the scope of the proposed Continual-MEGA benchmark extends beyond IAD, including a diverse range of anomalies and object types, similar to the variety found in widely used datasets such as MVTec-AD\mbox{~\citep{bergmann2019mvtec}}.
The ContinualAD dataset consists of a total of 30 classes, comprising 14,655 normal images and 15,827 anomaly images, notably larger than widely used MVTec-AD and VisA datasets.


\subsection{Benchmark Configuration}

\paragraph{Datasets Composition.}
{We compose a new benchmark to evaluate continual learning for anomaly detection in large-scale real-world settings, comprising various public datasets including MVTec-AD~\citep{bergmann2019mvtec}, VisA~\citep{zou2022spot}, Real-IAD~\citep{wang2024real}, VIADUCT~\citep{lehr2024ad3}, BTAD~\citep{mishra2021vt}, and MPDD~\citep{jezek2021deep}, and the newly proposed ContinualAD dataset, which consists of diverse images collected from real-world objects.
Figure~\ref{fig:example_images} shows example images from the seven datasets included in the proposed Continual-MEGA benchmark.
These examples highlight the diversity and complexity of anomaly types across domains.
To evaluate the continual learning performance, we design two experimental scenarios. The model is initially pre-trained on either 85 or 58 classes, followed by continual learning with 60 novel classes introduced incrementally.
}

\begin{figure*}[t]
\centering
\includegraphics[width=0.72\linewidth]{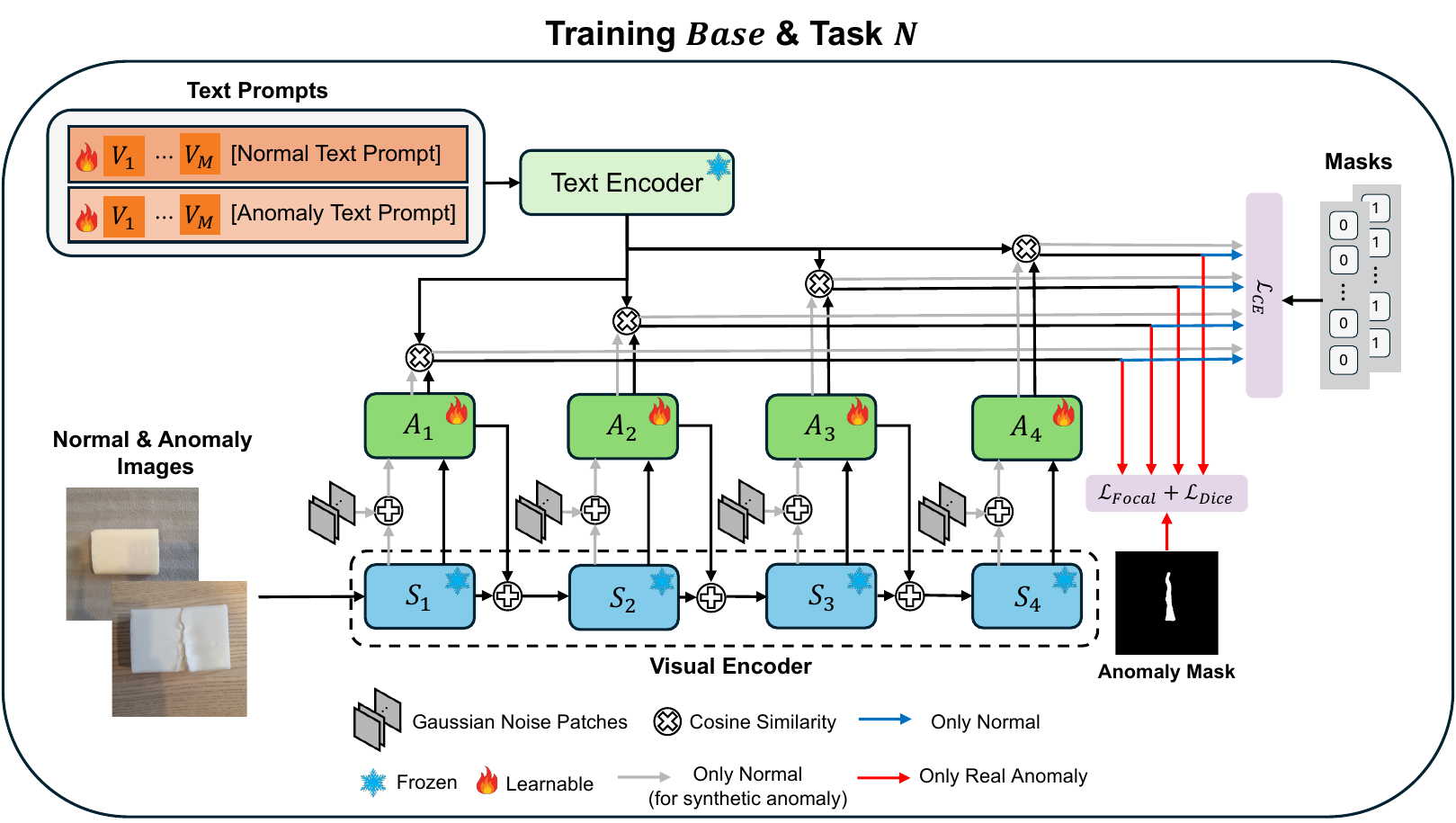}
\caption{\textbf{Overview architecture of training process.} Synthetic feature generation is performed exclusively during the training phase. At inference, adapters trained for each task are accumulated for evaluation. A detailed overview of the inference stage is provided in Appendix Figure~\ref{fig:overview_inference}.}
\vspace{-1.2em}
\label{fig:overview_training}
\end{figure*}

\paragraph{Various Scenarios for Continual Learning.}
To construct a large-scale continual learning benchmark for anomaly detection, we integrate seven datasets into three distinct evaluation scenarios. In each scenario, the continual learning setup is denoted as $(\#\text{Base})$-$(\#\text{New})$, where $(\#\text{Base})$ and $(\#\text{New})$ represent the number of base and newly introduced classes, referred to as \textit{Base} and \textit{New}, respectively.
The first two scenarios, \textbf{Scenario 1} and \textbf{Scenario 2}, represent the main evaluation of the benchmark. Furthermore, to evaluate the effectiveness of the proposed ContinualAD dataset, we conduct \textbf{Scenario 3} to compare it with \textbf{Scenario 2}.

\textbf{Scenario 1} extends conventional continual learning settings by combining the MVTec-AD~\citep{bergmann2019mvtec} and VisA~\citep{zou2022spot} datasets, widely used in anomaly detection. We pretrain the model on all $85$ \textit{Base} classes and sequentially introduce $5$, $10$, and $30$ \textit{New} classes over $12$, $6$, and $2$ iterations, respectively.
\textbf{Scenario 2} is designed to evaluate zero-shot generalization following continual adaptation. In this setting, both MVTec-AD and VisA are excluded from the continual learning process, as they are neither part of the \textit{Base} nor \textit{New} classes. Instead, they are held out solely for assessing the model's zero-shot performance, serving as a novel protocol to evaluate cross-domain generalization.
Additionally, \textbf{Scenario 3} further analyzes the generalization capability of the proposed \textit{ContinualAD} dataset by removing the target dataset from the \textit{Base} classes and \textit{New} classes stream. Specifically, the model is continually adapted with $30$ \textit{New} classes from other datasets, while zero-shot generalization is evaluated on the excluded dataset, following the setup of Scenario 2.
An overview of the three scenarios, including the base and new classes and the held-out datasets, is provided in Table~\ref{tab:scenario_overview} of the Appendix.

Detailed class distributions for Scenarios 1, 2, and 3 are illustrated in Appendix Figures~\ref{fig:scenario1_full},~\ref{fig:scenario2_full}, and~\ref{fig:scenario3_full}, respectively. These figures reveal a significant class imbalance across the datasets. Considering that we measure the amount of forgetting class-wise, we hypothesize that better fitting smaller classes from previous datasets~\citep{bergmann2019mvtec,zou2022spot} could be advantageous. We further investigate this supposition by comparing the results across the three scenarios in the Experiments section.

\paragraph{Metrics.}
For all quantitative evaluations, we adopt two metrics proposed in~\citep{tang2024incremental} for evaluating continual learning performance in anomaly detection: average accuracy (ACC) and forgetting measure (FM). The ACC metrics are computed on the basis of the image-level area under the ROC curve (AUROC) and the pixel-level average precision (AP)~\citep{liu2024unsupervised}, providing a comprehensive view of both the classification accuracy and the model's resilience to forgetting over time.
For the FM measure, we measure the decrease in ACC after adaptation.
To characterize the overall performance in more detail, we additionally report the mean ACC and FM over the image- and pixel-level scores, providing a compact summary of the joint performance across both granularities.

\paragraph{Training Protocol and Fair Comparison.}
All methods, including our proposed baseline, are trained under the same continual learning protocol in every scenario. 
Within each scenario, we use the same set of base classes and the same task order across all methods. 
For each method, the base classes are trained for 50 epochs and each continual adaptation phase is trained for 20 epochs, ensuring identical training budgets.
Following the concerns in~\citep{cha2024hyperparameters} about fair comparison in continual learning, we avoid per-scenario hyper-parameter tuning: for existing baselines, we use the publicly available default hyper-parameters and data augmentation settings from the authors’ implementations or papers, while for our proposed baseline method a single set of hyper-parameters is obtained via lightweight tuning on the base classes of Scenario~1 and then fixed for all remaining scenarios.
Moreover, to make the comparison more conservative, we do not apply any additional data augmentation for our baseline beyond the default preprocessing pipeline, whereas the compared methods use their standard data augmentation settings.
This setup enables a controlled and fair assessment of continual learning performance across all methods.

\section{Proposed Baseline Method}

To address the Continual-MEGA benchmark, we propose a simple baseline AD method, Anomaly Detection across Continual Tasks (ADCT), which leverages the text and visual encoders of CLIP~\citep{radford2021learning}, drawing inspiration from AnomalyCLIP~\citep{deng2022anomaly}. ADCT is designed to fully leverage CLIP information with minimal modification. 
ADCT employs a lightweight MLP \yj{adapter} for feature adaptation in each CLIP block, in conjunction with a feature synthesis module, design choices that may contribute to effective CZSL performance while preserving pre-trained CLIP knowledge.
Figure~\ref{fig:overview_training} illustrates the incremental training process on the $Base$ classes and successive Task $N$ classes, while Figure~\ref{fig:overview_inference} in the Appendix depicts the inference pathway formed by a mixture-of-experts adapters.

\begin{table*}[t]
\centering
\resizebox{0.9\textwidth}{!}{
\begin{tabular}{c|c|cc|cc|cc}
\Xhline{2\arrayrulewidth}
\multirow{2}{*}{Type}& \multirow{2}{*}{Method} & \multicolumn{2}{c|}{85-5 (12 tasks)} & \multicolumn{2}{c|}{85-10 (6 tasks)} & \multicolumn{2}{c}{85-30 (2 tasks)} \\
&& ACC($\uparrow$) & FM($\downarrow$) & ACC($\uparrow$) & FM($\downarrow$) & ACC($\uparrow$) & FM($\downarrow$) \\ \hline
\multirow{4}{*}{Only-normal} &SimpleNet   & 56.5/4.0/30.3 & 7.1/2.7/4.9 & 56.4/4.3/30.4 & 6.2/2.4/4.3 & 58.2/4.5/31.4 & 2.4/1.8/2.1 \\
&GeneralAD  & 49.3/1.5/25.4 & 5.5/1.2/3.4 & 50.2/1.4/25.8 & 3.2/1.5/2.4 & 48.9/1.1/25.0 & 5.8/1.2/3.5 \\
&HGAD    & 54.1/5.2/29.7 & 1.5/0.4/1.0 & 53.3/5.3/29.3 & 2.1/0.3/1.2 & 52.7/5.3/29.0 & 4.8/0.0/2.4 \\
&ResAD   & 73.1/13.9/43.5 & 1.3/0.4/0.8 & 71.9/12.7/42.3 & 1.0/0.3/0.6 & 70.3/10.1/40.2 & 0.2/1.5/0.8 \\ \hline
\multirow{4}{*}{VLM-based} &MVFA   & 75.4/24.4/\textbf{49.9} & 4.2/5.6/4.9 & 76.4/24.3/50.4 & 4.0/6.8/5.4 & 75.7/24.8/50.3 & 6.3/10.3/8.3 \\
&AnomalyCLIP & 48.8/2.7/25.8 & 9.2/1.1/5.2 & 50.6/2.5/26.6 & 4.5/0.9/2.7 & 50.5/2.6/26.6 & 2.3/0.3/1.3 \\
&VCP-CLIP & 44.1/19.3/31.7 & 2.5/9.0/5.7 & 61.9/25.6/43.7 & 4.4/4.1/4.2 & 44.7/28.9/36.8 & 4.0/2.4/3.2 \\
&MediCLIP & \textbf{80.5}/8.8/44.7 & 1.4/6.0/3.7 & \textbf{77.9}/6.9/42.4 & 2.2/10.2/6.2 & 77.7/9.7/43.7 & 0.4/20.0/10.2 \\ \hline
\multirow{4}{*}{Continual} &UCAD  & 67.1/10.8/39.0 & 0.2/0.0/0.1 & 64.6/7.8/36.2 & 0.3/0.03/0.2 & 57.9/4.4/31.2 & 1.2/0.0/0.6 \\
&IUF & 59.8/5.8/32.8 & 1.3/0.3/0.8 & 60.1/6.0/33.1 & 0.1/0.1/0.1 & 59.8/5.9/32.9 & 0.5/0.4/0.5 \\
&$\text{IUF}^\ast$ & 61.5/7.4/34.5 & 0.5/0.3/0.4 & 61.4/7.6/34.5 & 0.5/0.1/0.3 & 63.0/8.8/35.9 & 0.4/0.3/0.4 \\
&\textbf{Ours} & 73.8/\textbf{25.7}/49.8 & 2.0/2.1/2.1 & 75.8/\textbf{28.0}/\textbf{51.9} & 1.3/1.9/1.6 & \textbf{78.9}/\textbf{32.7}/\textbf{55.8} & 0.8/1.8/1.3 \\

\Xhline{2\arrayrulewidth}
\end{tabular}
}
\caption{\textbf{Experimental results on Scenario 1.} $\cdot/\cdot/\cdot$ denotes the Image-AUROC, Pixel-AP and their average, respectively. While all methods were trained with the same number of epochs for fair comparison, the $\text{IUF}^\ast$ method requires substantially longer training due to its methodology. Thus, we trained the $Base$ classes for 500 epochs and the $New$ classes for 100 epochs in the $\text{IUF}^\ast$ setting. 
The notation ``$X$-$Y$ ($Z$ tasks)'' in the first row denotes an evaluation setup where the model is initially trained on $X$ base classes, followed by $Z$ continual learning phases (tasks), each comprising $Y$ new classes.
}
\vspace{-1.5em}
\label{tab:scenario1}
\end{table*}

\subsection{Mixture-of-Experts of Adapters}
To design the adapter, we \yj{employ a set of four adapters} $A = \{A_1, \dots A_4\}$ for each block of layers of the CLIP visual encoder. For each category including the set of (\textit{Base}) classes $C_b$ and the set of $n$'th task $C_n$ classes, where the total number of tasks is $n=1,\dots N$, we separately train the adapter set denoted as $A_n = \{A_{1,n}, \dots, A_{4,n}\}$. 

In the inference stage, as illustrated in Appendix Figure~\ref{fig:overview_inference}, we accumulate all the pre-trained adapters $A_n$ and $A_b$ for each task and \textit{Base} classes set, by the average adapters $A^{\text{avg}} = \{A^{\text{avg}}_{1},..., A^{\text{avg}}_{4}\}$, as follows:

\begin{eqnarray}
\begin{aligned}
\label{eq:synthesis_function}
A^{\text{avg}}_{l} = \frac{1}{N+1}(\sum^N_n {A_{l,n}} + A_{l,b}).
\end{aligned}
\end{eqnarray}
where the number $l=\{1,\dots 4\}$ denotes the index of each of the four blocks.
In implementation, each adaptation layer $A_l(\cdot)$ consists of two linear layers as:
\begin{equation}
    A_l(F_l) = W_{l,2}(W_{l,1}F_l^T),
\end{equation}
where $F_l$ represents the visual features extracted from the $l$-th visual encoder stage of CLIP.
We use CLIP with ViT-L/14~\citep{dosovitskiy2020image} architecture, which consists of 24 sublayers divided into four layers, where each layer contains six sublayers. The size of input images was set to 336. The adaptation layers for anomaly feature generation were applied to layers 1, 2, 3, and 4.

\subsection{Synthetic Feature Generation}
Specifically, we apply random noise to enable the adaptation layers $A_l$ to learn a diverse range of anomalies. In training, we use task-wise adapters $A_n$ and in the inference phase, we use the accumulated adapter $A^{\text{avg}}$.
The synthetic anomaly features ($F_l^1$) are generated by 
\begin{equation}
    F_l^1 = A_l(F_l + \gamma),
    \label{eq:random}
\end{equation}
where $\gamma \in \mathbb{R}^{G \times d}$ is random noise.
The adapted normal features ($F_l^0$) are generated via the adaptation layers as
\begin{equation}
    F_l^0 = A_l(F_l).
\end{equation}
The adapted normal features $F_l^0$ and synthetic anomaly features $F_l^1$ are both used to generate anomaly score maps by calculating cosine similarity along with the text features.

\section{Experiments}
\label{sec:experiment}

\begin{table*}[t]
\centering
\resizebox{\textwidth}{!}{
\begin{tabular}{l|l|cc|cc|cc|cc}
\Xhline{2\arrayrulewidth}
\multirow{2}{*}{Type}& \multirow{2}{*}{Method} & \multicolumn{2}{c|}{58-5 (12 tasks)} & \multicolumn{2}{c|}{58-10 (6 tasks)} & \multicolumn{2}{c|}{58-30 (2 tasks)} & \multicolumn{2}{c}{zero-shot (Avg.)} \\ 
&& ACC($\uparrow$) & FM($\downarrow$) & ACC($\uparrow$) & FM($\downarrow$) & ACC($\uparrow$) & FM($\downarrow$) & MVTec-AD & VisA  \\ \hline
\multirow{4}{*}{Only-normal} &SimpleNet  & 56.1/4.2/30.2 & 8.2/1.4/4.8 & 56.5/3.8/30.2 & 7.1/2.4/4.8 & 57.3/3.8/30.6 & 4.9/1.0/3.0 & 55.8/9.7/32.8 & 52.6/0.0/26.3  \\
&GeneralAD  & 49.0/0.8/24.9 & 6.3/1.7/4.0 & 51.3/0.9/26.1 & 3.1/1.2/2.2 & 47.7/2.1/24.9 & 5.7/0.0/2.9   & 53.3/5.8/29.6 & 49.2/1.4/25.3\\
&HGAD   & 51.1/4.3/27.7 & 1.8/0.3/1.1 & 51.8/4.5/28.2 & 1.4/0.2/0.8 & 51.8/4.3/28.1 & 2.4/0.2/1.3  & 50.1/16.1/33.1 & 55.1/2.7/28.9 \\
&ResAD   & 48.8/0.6/24.7 & 10.7/1.0/5.8 & 42.7/1.7/22.2 & 3.0/0.9/1.9 & 55.6/12.8/34.2 & 12.0/4.7/8.3  & 69.7/11.1/40.4 & 57.8/3.1/30.4 \\ \hline
\multirow{4}{*}{VLM-based} &MVFA   & 63.2/4.7/34.0 & 5.8/5.4/5.6 & 64.0/4.1/34.1 & 5.8/2.9/4.4 & 65.3/5.0/35.2 & 1.9/2.0/2.0  & 56.1/5.1/30.6 & 53.8/2.5/28.2 \\
 &AnomalyCLIP  & 52.9/2.0/27.5 & 4.1/0.9/2.5 & 51.3/1.9/26.6 & 1.5/0.6/1.1 & 51.1/2.2/26.7 & 2.2/0.2/1.2 & 57.2/7.0/32.1 & 51.3/3.6/27.5 \\
&VCP-CLIP & 55.6/18.7/37.1 & 3.8/6.8/5.3 & 53.2/19.8/36.5 & 0.3/3.0/1.7 & 64.3/22.3/48.3 & 2.8/3.7/3.3  & 62.3/22.7/42.5 & 61.0/11.2/36.1 \\
&MediCLIP     & \textbf{79.6}/7.3/43.5 & 3.8/5.6/4.7 & \textbf{76.0}/6.0/41.0 & 4.9/3.7/4.3 & \textbf{77.1}/5.9/41.5 & 2.1/7.0/4.6  & \textbf{84.2}/19.1/51.7 & 74.1/5.2/39.7\\ \hline
\multirow{4}{*}{Continual} & UCAD  & 66.0/7.4/36.7 & 0.4/0.02/0.2 & 63.5/6.0/34.8 & 0.7/0.03/0.4 & 58.0/3.1/30.6 & 0.0/0.0/0.0  & 61.6/9.4/35.5 & 54.1/1.9/28.0\\
& IUF & 57.6/4.2/30.9 & 1.7/0.5/1.1 & 58.0/4.3/31.2 & 0.3/0.2/0.3 & 58.0/4.3/31.2 & -0.7/-0.1/-0.4  & 68.0/16.2/42.1 & 54.7/2.8/28.8\\
& $\text{IUF}^\ast$  & 60.2/6.3/33.3 & 0.8/0.3/0.6 & 60.7/6.4/33.6 & 0.2/0.1/0.2 & 61.7/7.0/34.4 & 0.2/0.2/0.2  & 67.8/15.4/41.6 & 58.2/4.9/31.6\\ 
& Ours & 71.7/\textbf{20.7}/\textbf{46.2} & 2.3/4.1/3.2 & 72.4/\textbf{22.2}/\textbf{47.3} & 2.5/3.8/3.2 & 76.8/\textbf{27.5}/\textbf{52.2} & 1.0/2.6/1.8  & 78.4/\textbf{31.5}/\textbf{55.0} & \textbf{76.9}/\textbf{17.2}/\textbf{47.0}\\

\Xhline{2\arrayrulewidth}
\end{tabular}
}
\caption{\textbf{Experimental results on Scenario 3.} To verify the effectiveness of the ContinualAD dataset, it is excluded from the training process. $\cdot/\cdot/\cdot$ denotes the Image-AUROC, Pixel-AP, and their average, respectively. The notation ``$X$-$Y$ ($Z$ tasks)'' denotes an evaluation setup where the model is initially trained on $X$ base classes, followed by $Z$ continual learning phases (tasks), each comprising $Y$ new classes.}
\vspace{-1.0em}
\label{tab:scenario2_full}
\end{table*}

\paragraph{Overview.}
Tables~\ref{tab:scenario1} and \ref{tab:scenario2_full} present the quantitative results under the proposed evaluation scenarios, comparing various recently proposed anomaly detection methods. In our experiments, we categorize the methods into three groups: (1) approaches that adapt using only normal samples: SimpleNet~\citep{liu2023simplenet}, GeneralAD~\citep{strater2024generalad}, HGAD~\citep{yao2024hierarchical}, and ResAD~\citep{yao2024resad}; (2) vision-language model (VLM)-based methods: MVFA~\citep{huang2024adapting}, VCP-CLIP~\citep{qu2024vcp}, and MediCLIP~\citep{zhang2024mediclip}; and (3) methods specifically designed for continual learning settings: UCAD~\citep{liu2024unsupervised} and IUF~\citep{tang2024incremental}.
Overall, the results indicate a substantial drop in performance across all methods, particularly for pixel-wise anomaly detection, when evaluated under the proposed continual learning settings. This contrasts sharply with the higher performance typically observed in standard benchmarks such as MVTec-AD and VisA, highlighting both the increased difficulty and practical relevance of our evaluation.

\begin{table*}[t]
\centering
\resizebox{\textwidth}{!}{
\begin{tabular}{l|l|cc|cc|cc|cc}
\Xhline{2\arrayrulewidth}
\multirow{2}{*}{Type}& \multirow{2}{*}{Method} & \multicolumn{2}{c|}{58-5 (6 tasks)} & \multicolumn{2}{c|}{58-10 (3 tasks)} & \multicolumn{2}{c|}{58-30 (1 tasks)} & \multicolumn{2}{c}{zero-shot (Avg.)} \\ 
&& ACC($\uparrow$) & FM($\downarrow$) & ACC($\uparrow$) & FM($\downarrow$) & ACC($\uparrow$) & FM($\downarrow$) & MVTec-AD & VisA  \\ \hline
\multirow{4}{*}{Only-normal} 
&SimpleNet      & 57.6/5.5/31.6 & 7.2/3.0/5.1 & 59.5/7.2/33.4 & 5.9/2.4/4.2 & 59.8/6.8/33.3 & 2.2/0.4/1.3 & 50.2/7.8/29.0 & 49.9/0.0/25.0  \\
&GeneralAD      & 50.6/0.7/25.7 & 3.3/2.0/2.7 & 51.7/1.0/26.3 & 5.3/2.6/3.9 & 51.7/1.4/26.6 & 3.3/0.9/2.1 & 52.0/6.3/29.2 & 51.7/2.5/27.1  \\
&HGAD           & 53.2/3.7/28.5 & 2.5/0.1/1.3 & 53.2/3.8/28.5 & 2.9/0.0/1.4 & 53.4/3.7/28.6 & 2.7/0.0/1.4 & 49.5/15.6/32.6 & 57.4/2.9/30.2  \\
&ResAD          & 44.6/2.3/23.5 & 1.3/0.3/0.8 & 40.5/0.8/20.7 & 7.9/4.3/6.1 & 64.2/4.1/34.2 & 7.8/0.8/4.3 & 78.0/12.1/45.1 & 67.6/7.6/37.6  \\ \hline
\multirow{4}{*}{VLM-based} 
&MVFA           & 63.3/6.2/34.8 & 8.1/10.7/9.4 & 68.0/11.0/39.5 & 3.8/10.1/7.0 & 69.6/16.4/43.0 & 3.7/5.4/4.6 & 69.7/9.8/39.8 & 69.8/5.3/37.6  \\
&AnomalyCLIP     & 51.4/2.6/27.0 & 5.3/1.7/3.5 & 53.5/2.7/28.1 & 1.3/0.4/0.9 & 54.1/3.1/28.6 & -1.1/0.2/-0.4 & 51.7/6.8/29.3 & 49.9/2.7/26.3  \\
&VCP-CLIP       & 54.3/\textbf{21.2}/37.8 & 1.9/2.9/2.4 & 46.1/18.1/32.1 & -0.2/3.3/1.6 & 61.5/21.0/41.3 & 2.1/1.2/1.6 & 58.9/22.5/40.7 & 58.0/10.6/34.3 \\
&MediCLIP       & \textbf{77.3}/7.1/42.2 & 3.6/4.6/4.1 & \textbf{76.0}/5.0/40.5 & 2.0/2.9/2.5 & 73.2/5.3/39.3 & 6.3/3.5/4.9 & \textbf{81.8}/17.3/49.6 & \textbf{74.7}/4.3/39.5  \\ \hline
\multirow{4}{*}{Continual} 
&UCAD           & 65.0/9.6/37.3 & 0.0/0.0/0.0 & 59.8/5.8/32.8 & 0.0/0.0/0.0 & 55.2/3.4/29.3 & 0.0/0.0/0.0 & 59.7/9.0/34.3 & 53.6/1.7/27.7 \\
&IUF            & 58.1/4.6/31.4 & 1.2/0.4/0.8 & 57.6/4.4/31.0 & 0.1/0.2/0.2 & 57.6/4.2/30.9 & 0.3/0.1/0.2 & 67.8/15.8/41.8 & 54.9/2.7/28.8  \\ 
&$\text{IUF}^\ast$ & 59.3/6.3/32.8 & 0.5/0.3/0.4 & 59.7/6.5/33.1 & 0.8/0.1/0.5 & 60.9/7.4/34.2 & 0.5/0.4/0.5 & 64.6/14.5/39.6 & 57.8/3.5/30.7  \\ 
&\textbf{Ours}  & 69.5/19.7/\textbf{44.6} & 3.2/3.4/3.3 & 72.7/\textbf{23.1}/\textbf{47.9} & 2.4/3.7/3.1 & \textbf{76.8}/\textbf{29.5}/\textbf{53.2} & -0.3/2.1/0.9 & 75.0/\textbf{28.4}/\textbf{51.7} & 69.7/\textbf{13.7}/\textbf{41.7} \\

\Xhline{2\arrayrulewidth}
\end{tabular}
}
\caption{\textbf{Experimental results on Scenario 3.} To verify the effectiveness of the ContinualAD dataset, it is excluded from the training process. $\cdot/\cdot/\cdot$ denotes the Image-AUROC, Pixel-AP, and their average, respectively. The notation ``$X$-$Y$ ($Z$ tasks)'' denotes an evaluation setup where the model is initially trained on $X$ base classes, followed by $Z$ continual learning phases (tasks), each comprising $Y$ new classes.}
\vspace{-1.5em}
\label{tab:scenario3}
\end{table*}


\begin{figure}[t]
\centering
    \centering
    \includegraphics[width=\linewidth]{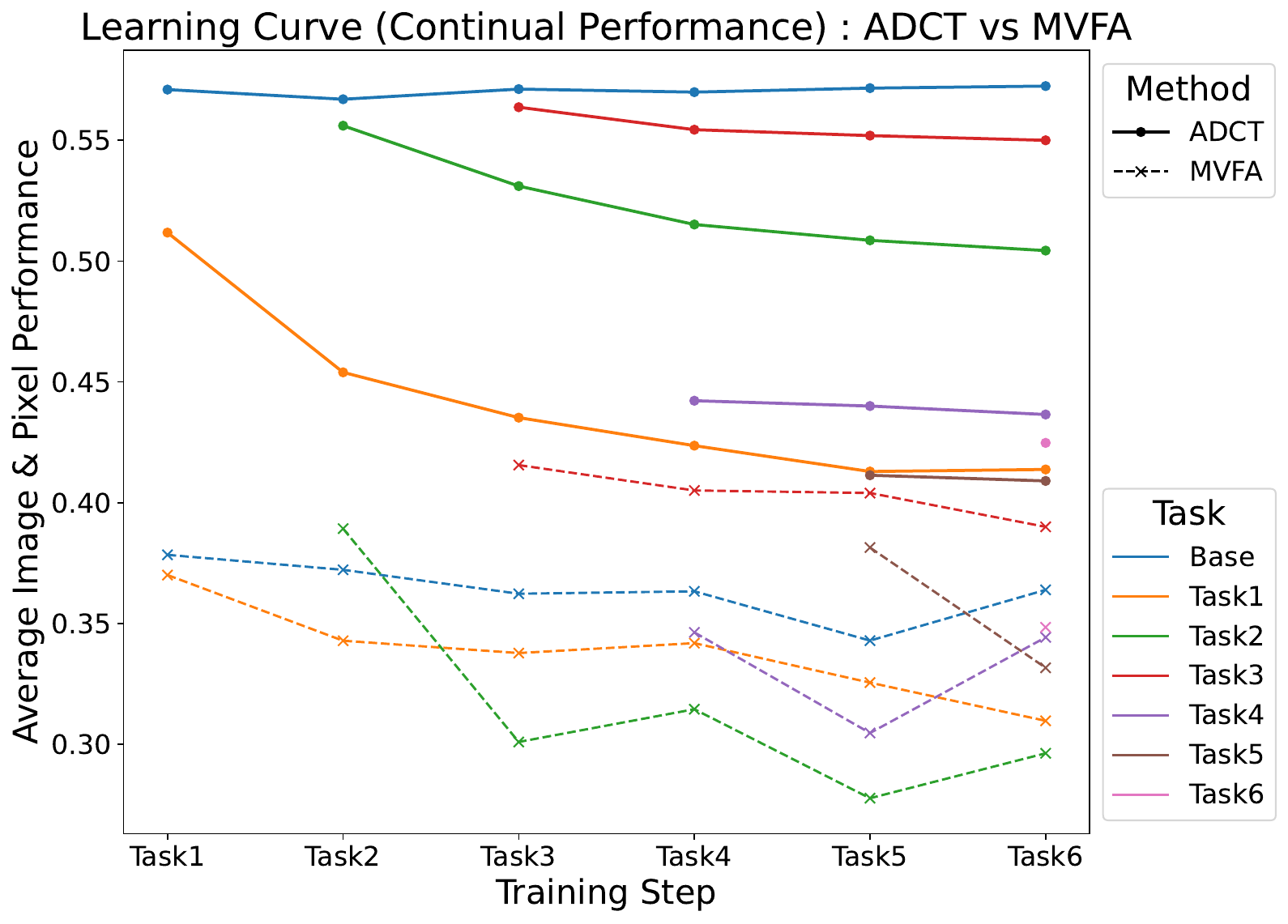}
    \caption{\textbf{Learning curves on Scenario~2 (10 classes per task).}
    We plot the average of image-level AUROC and pixel-level AP over all tasks seen so far as the model is incrementally trained from Task~1 to Task~6.
This plot highlights that under matched compute, ADCT (Ours) exhibits superior stability, preserving performance across tasks substantially better than MVFA, which corroborates our quantitative FM results in Table~\ref{tab:scenario2_full}.
    }
    \label{fig:learning_curve}
    \vspace{-2.0em}
\end{figure}

\begin{table*}[t]
\centering
\resizebox{0.9\textwidth}{!}{
\begin{tabular}{ccc|c|c|c|c|c|c|c|c}
\Xhline{2\arrayrulewidth}
\multicolumn{3}{c|}{Components}         & \multicolumn{4}{c|}{58-30 (2 tasks)}  & \multicolumn{4}{c}{Zero-shot} \\ \hline
\multirow{2}{*}{\emph{Adapters}} & \multirow{2}{*}{\emph{Synthetic}}  & \multirow{2}{*}{\emph{Mixture}}       & \multicolumn{2}{c|}{Image}      & \multicolumn{2}{c|}{Pixel} & \multicolumn{2}{c|}{MVTec-AD}      & \multicolumn{2}{c}{VisA}    \\ \cline{4-11}
& &                                      & ACC  & FM   & ACC & FM  & Image  & Pixel   & Image & Pixel            \\
\hline
           &            &                & 56.0 &  --  & 1.0 & -- & 75.2 & 2.3  & 61.8 & 1.0 \\
\checkmark &            & \checkmark     & $77.7 \pm 0.6$ & $0.6 \pm 0.4$ & $22.3 \pm 3.1$ & $0.3 \pm 0.3$   & $86.9 \pm 1.0$ & $26.3 \pm 4.0$ & $82.8 \pm 0.1$ & $13.9 \pm 3.8$\\ 
\checkmark & \checkmark &                & $77.2 \pm 0.5$ & $4.1 \pm 0.2$ & $30.1 \pm 1.1$ & $6.7 \pm 1.7$   & $82.5 \pm 1.2$ & $35.7 \pm 0.2$ & $77.7 \pm 3.3$ & $19.7 \pm 0.7$ \\       
\checkmark & \checkmark & \checkmark     & $76.3 \pm 0.4$ & $1.4 \pm 0.5$ & $26.8 \pm 0.8$ & $2.8 \pm 0.2$ & $81.2 \pm 0.8$ & $32.1 \pm 0.8$ & $78.8 \pm 0.2$ & $18.8 \pm 0.3 $\\                

\Xhline{2\arrayrulewidth}
\end{tabular}
}
\caption{\textbf{Ablation study} on Scenario 2. All results except the first row are averaged over three random seeds and reported as mean $\pm$ standard deviation. 
The first row without any components corresponds to vanilla pretrained CLIP without any continual learning; thus FM is not defined (shown as ``--''), and the reported scores are single values rather than mean $\pm$ standard deviation.}
\vspace{-1.5em}
\label{tab:ablation}
\end{table*}

\paragraph{Evaluation of Continual Learning Capability. }
Scenario 1 represents a typical continual learning setup but significantly scales up both the number of classes and the volume of data compared to prior works on continual adaptation~\citep{liu2024unsupervised,tang2024incremental}. 

Notably, vision-language model (VLM)-based methods such as MVFA~\citep{huang2024adapting} and MediCLIP~\citep{zhang2024mediclip} achieve the highest performance among all baselines, apart from the proposed method. Specifically, MVFA and our proposed method show comparable performance. 
The overall results strongly suggest that existing methods, including continual learning approaches for anomaly detection, struggle to handle large-scale continual evaluation settings.
This observation reveals two key insights: (1) several prior methods appear to be tightly fitted to existing benchmarks such as MVTec-AD and VisA, which limits their generalizability to more diverse or challenging settings; and (2) methods with stronger initial (pretrained) performance tend to retain higher accuracy throughout continual adaptation.

Regarding the first insight, methods such as MVFA, which demonstrate competitive performance in Scenario~1, exhibit significantly degraded results, particularly in pixel-level AP, when MVTec-AD and VisA datasets are excluded from the \textit{Base} and \textit{New} classes, as shown in Table~\ref{tab:scenario2_full}. 
In contrast, our proposed method consistently achieves robust performance across all evaluation scenarios.
Regarding the second insight, VLM-based methods demonstrate significantly stronger performance compared to methods explicitly designed for continual learning. This discrepancy can be attributed to the limited detection capability of existing continual anomaly detection methods, even at their initial stage. 
Consequently, these methods face greater difficulty in adapting to new incoming categories.
Although the forgetting measure (FM) of continual learning-based methods appears lower than that of VLM-based methods, this is likely due to their poor initial detection performance rather than effective forgetting mitigation.

\paragraph{Evaluation of Generalizability and Stability.}
Another notable aspect of the evaluation is the CZSL scenario, which evaluates zero-shot generalization following continual adaptation.
In Scenario 2, our proposed baseline model demonstrates improved generalization, benefiting from both a stronger set of \textit{Base} classes and the continual adaptation of additional classes. Scenario 2 includes our proposed ContinualAD dataset, under which most methods exhibit improved zero-shot performance after continual learning. In contrast, when ContinualAD is excluded in Scenario 3, most existing methods suffer a degradation in zero-shot generalization after continual learning. Among them, our proposed method shows the smallest performance drop, indicating stronger robustness to continual adaptation compared to other approaches.
To further investigate this observation, we refer to the quantitative results from Scenarios 2 and 3, presented in Table~\ref{tab:scenario2_full} and Table~\ref{tab:scenario3}. These tables provide detailed evaluation metrics corresponding to continual adaptation and zero-shot evaluation results.
\yj{To facilitate comparison}, we visualize the results for representative methods of the tables in Figure~\ref{fig:base_continual_zero} of the Appendix.

To further analyze how performance evolves as new tasks arrive,
Figure~\ref{fig:learning_curve} presents the learning curves in Scenario~2 with 10 classes per task and six tasks in total.
For each increment (Task~1~$\rightarrow$~Task~6), we plot the average of image-level AUROC and pixel-level AP over all tasks observed so far, under a matched-compute setting (identical epochs, input resolution, and continual stream).
Across all increments, MVFA consistently underperforms ADCT and exhibits a sharper degradation as new tasks are introduced, indicating more pronounced forgetting of earlier tasks, whereas ADCT maintains a higher and flatter curve, demonstrating superior stability.
These observations are consistent with our FM results and underscore that explicitly controlling forgetting is crucial for stable continual anomaly detection.

Notably, we also observe that methods such as MediCLIP~\citep{zhang2024mediclip} and our proposed ADCT, both leveraging CLIP with lightweight adapters and image or feature synthesis, consistently achieve competitive performance in the CZSL scenario.
Based on these results, we can conjecture that \textit{our ContinualAD dataset provides valuable information for detecting anomalies in both unseen and continually introduced categories under more challenging scenarios than prior setups. Our baseline demonstrates robust and consistent performance, suggesting a potentially valuable insight: lightweight CLIP adaptation combined with feature synthesis may contribute to improved continual and CZSL learning.}

\paragraph{Ablation Study.}
Table~\ref{tab:ablation} presents an ablation on Scenario~2 (30 classes per task), isolating three components: the adapter-based continual learner (\emph{Adapters}), synthetic anomaly generation (\emph{Synthetic}), and the mixture-of-expert adapters inference scheme (\emph{Mixture}).  
The first row corresponds to vanilla CLIP without any of these components, so FM in the continual setting is undefined and omitted. 
All configurations with at least one component outperform vanilla CLIP in both the continual setting and the zero-shot evaluations. 
Comparing rows with and without \emph{Synthetic}, we observe that omitting synthetic anomaly generation (\emph{Adapters} + \emph{Mixture}) leads to noticeably lower pixel-level scores in both the continual setting and the zero-shot evaluation, highlighting that synthetic anomalies are crucial for strong pixel-level performance.
Since image-level metrics can be overly optimistic when anomaly localization is poor, we view pixel-level scores as a more reliable indicator of anomaly detection quality.
Using \emph{Adapters} with \emph{Synthetic} but without \emph{Mixture} (\emph{Adapters} + \emph{Synthetic}) gives high overall performance but substantially larger FM, indicating strong forgetting of previous tasks.
In contrast, the full configuration (\emph{Adapters} + \emph{Synthetic} + \emph{Mixture}) maintains strong zero-shot performance while significantly reducing FM, which is desirable for a continual-learning baseline that aims to minimize forgetting without sacrificing zero-shot accuracy.
\section{Conclusion}

We present Continual-MEGA, a new large-scale benchmark for continual anomaly detection (AD), constructed by integrating multiple public datasets and curating a novel dataset, ContinualAD, to significantly enhance sample volume and diversity. Comprehensive evaluations on Continual-MEGA reveal that existing AD methods still have substantial room for improvement, highlighting both the need for further research in continual AD and the effectiveness of the curated ContinualAD dataset.
\gu{To address these challenges}, we introduce a novel baseline method integrating MoE-style adapter modules, anomaly feature synthesis, and \gu{prompt tuning within the CLIP framework.} 
Our baseline exhibits robust and consistent performance, \gu{highlighting} a promising direction: \gu{the synergy between} lightweight CLIP adaptation and feature \gu{synthesis} \gu{is vital for achieving both learning stability and zero-shot generalization within continual frameworks.}




\section*{Impact Statement}

This research contributes to the field of industrial anomaly detection by establishing a large-scale benchmark for continual and zero-shot learning. A primary motivation of our work is to mitigate the computational cost and carbon footprint associated with retraining deep learning models in dynamic production environments. The proposed datasets and methods focus exclusively on product defects and do not involve personal identifiable information or surveillance capabilities. Consequently, we believe this work presents minimal ethical risks while offering tangible benefits for manufacturing sustainability and reproducibility in machine learning research.


\bibliography{icml_paper}

@article{dosovitskiy2020image,
  title={An image is worth 16x16 words: Transformers for image recognition at scale},
  author={Dosovitskiy, Alexey and Beyer, Lucas and Kolesnikov, Alexander and Weissenborn, Dirk and Zhai, Xiaohua and Unterthiner, Thomas and Dehghani, Mostafa and Minderer, Matthias and Heigold, Georg and Gelly, Sylvain and others},
  journal={arXiv preprint arXiv:2010.11929},
  year={2020}
}

@inproceedings{zhang2023continual,
  title={Continual zero-shot learning through semantically guided generative random walks},
  author={Zhang, Wenxuan and Janson, Paul and Yi, Kai and Skorokhodov, Ivan and Elhoseiny, Mohamed},
  booktitle={Proceedings of the IEEE/CVF international conference on computer vision},
  pages={11574--11585},
  year={2023}
}

@article{skorokhodov2020class,
  title={Class normalization for (continual)? generalized zero-shot learning},
  author={Skorokhodov, Ivan and Elhoseiny, Mohamed},
  journal={arXiv preprint arXiv:2006.11328},
  year={2020}
}

@inproceedings{zhang2024continual,
  title={Continual compositional zero-shot learning},
  author={Zhang, Yang and Feng, Songhe and Yuan, Jiazheng},
  booktitle={Proceedings of the Thirty-Third International Joint Conference on Artificial Intelligence},
  pages={1724--1732},
  year={2024}
}

@incollection{gautam2024generative,
  title={Generative replay-based continual zero-shot learning},
  author={Gautam, Chandan and Parameswaran, Sethupathy and Mishra, Ashish and Sundaram, Suresh},
  booktitle={Towards Human Brain Inspired Lifelong Learning},
  pages={73--100},
  year={2024},
  publisher={World Scientific}
}

@article{chaudhry2018efficient,
  title={Efficient lifelong learning with a-gem},
  author={Chaudhry, Arslan and Ranzato, Marc'Aurelio and Rohrbach, Marcus and Elhoseiny, Mohamed},
  journal={arXiv preprint arXiv:1812.00420},
  year={2018}
}

@inproceedings{bergmann2019mvtec,
  title={MVTec AD--A comprehensive real-world dataset for unsupervised anomaly detection},
  author={Bergmann, P. and Fauser, M. and Sattlegger, D. and Steger, C.},
  booktitle={Proceedings of the IEEE/CVF Conference on Computer Vision and Pattern Recognition},
  pages={9592--9600},
  year={2019}
}

@article{loshchilov2017decoupled,
  title={Decoupled weight decay regularization},
  author={Loshchilov, I},
  journal={arXiv preprint arXiv:1711.05101},
  year={2017}
}

@article{cha2024hyperparameters,
  title={Hyperparameters in continual learning: a reality check},
  author={Cha, Sungmin and Cho, Kyunghyun},
  journal={arXiv preprint arXiv:2403.09066},
  year={2024}
}

@inproceedings{zou2022spot,
  title={Spot-the-difference self-supervised pre-training for anomaly detection and segmentation},
  author={Zou, Y. and Jeong, J. and Pemula, L. and Zhang, D. and Dabeer, O.},
  booktitle={European Conference on Computer Vision},
  pages={392--408},
  year={2022},
  publisher={Springer Nature Switzerland}
}

@inproceedings{jezek2021deep,
  title={Deep learning-based defect detection of metal parts: evaluating current methods in complex conditions},
  author={Jezek, S. and Jonak, M. and Burget, R. and Dvorak, P. and Skotak, M.},
  booktitle={2021 13th International Congress on Ultra Modern Telecommunications and Control Systems and Workshops (ICUMT)},
  pages={66--71},
  year={2021},
  publisher={IEEE}
}

@inproceedings{deng2022anomaly,
  title={Anomaly detection via reverse distillation from one-class embedding},
  author={Deng, H. and Li, X.},
  booktitle={Proceedings of the IEEE/CVF Conference on Computer Vision and Pattern Recognition},
  pages={9737--9746},
  year={2022}
}

@inproceedings{roth2022towards,
  title={Towards total recall in industrial anomaly detection},
  author={Roth, K. and Pemula, L. and Zepeda, J. and Sch{\"o}lkopf, B. and Brox, T. and Gehler, P.},
  booktitle={Proceedings of the IEEE/CVF Conference on Computer Vision and Pattern Recognition},
  pages={14318--14328},
  year={2022}
}

@inproceedings{defard2021padim,
  title={Padim: a patch distribution modeling framework for anomaly detection and localization},
  author={Defard, T. and Setkov, A. and Loesch, A. and Audigier, R.},
  booktitle={International Conference on Pattern Recognition},
  pages={475--489},
  year={2021},
}

@inproceedings{liu2023simplenet,
  title={Simplenet: A simple network for image anomaly detection and localization},
  author={Liu, Z. and Zhou, Y. and Xu, Y. and Wang, Z.},
  booktitle={Proceedings of the IEEE/CVF Conference on Computer Vision and Pattern Recognition},
  pages={20402--20411},
  year={2023}
}

@inproceedings{strater2024generalad,
  title={GeneralAD: Anomaly Detection Across Domains by Attending to Distorted Features},
  author={Str{\"a}ter, L. P. and Salehi, M. and Gavves, E. and Snoek, C. G. and Asano, Y. M.},
  booktitle={Proceedings of the European Conference on Computer Vision},
  year={2024}
}

@inproceedings{you2022unified,
  title={A unified model for multi-class anomaly detection},
  author={You, Z. and Cui, L. and Shen, Y. and Yang, K. and Lu, X. and Zheng, Y. and Le, X.},
  booktitle={Advances in Neural Information Processing Systems},
  volume={35},
  pages={4571--4584},
  year={2022}
}

@article{hendrycks2021unsolved,
  title={Unsolved problems in ml safety},
  author={Hendrycks, Dan and Carlini, Nicholas and Schulman, John and Steinhardt, Jacob},
  journal={arXiv preprint arXiv:2109.13916},
  year={2021}
}

@article{amodei2016concrete,
  title={Concrete problems in AI safety},
  author={Amodei, Dario and Olah, Chris and Steinhardt, Jacob and Christiano, Paul and Schulman, John and Man{\'e}, Dan},
  journal={arXiv preprint arXiv:1606.06565},
  year={2016}
}

@inproceedings{yao2024hierarchical,
  title={Hierarchical Gaussian Mixture Normalizing Flow Modeling for Unified Anomaly Detection},
  author={Yao, X. and Li, R. and Qian, Z. and Wang, L. and Zhang, C.},
  booktitle={Proceedings of the European Conference on Computer Vision},
  year={2024}
}

@inproceedings{jeong2023winclip,
  title={Winclip: Zero-/few-shot anomaly classification and segmentation},
  author={Jeong, J. and Zou, Y. and Kim, T. and Zhang, D. and Ravichandran, A. and Dabeer, O.},
  booktitle={Proceedings of the IEEE/CVF Conference on Computer Vision and Pattern Recognition},
  pages={19606--19616},
  year={2023}
}

@inproceedings{huang2024adapting,
  title={Adapting visual-language models for generalizable anomaly detection in medical images},
  author={Huang, C. and Jiang, A. and Feng, J. and Zhang, Y. and Wang, X. and Wang, Y.},
  booktitle={Proceedings of the IEEE/CVF Conference on Computer Vision and Pattern Recognition},
  pages={11375--11385},
  year={2024}
}

@inproceedings{cao2024adaclip,
  title={Adaclip: Adapting clip with hybrid learnable prompts for zero-shot anomaly detection},
  author={Cao, Y. and Zhang, J. and Frittoli, L. and Cheng, Y. and Shen, W. and Boracchi, G.},
  booktitle={Proceedings of the European Conference on Computer Vision},
  year={2024},
  month={July}
}

@inproceedings{zhou2023anomalyclip,
  title={Anomalyclip: Object-agnostic prompt learning for zero-shot anomaly detection},
  author={Zhou, Q. and Pang, G. and Tian, Y. and He, S. and Chen, J.},
  booktitle={Proceedings of the International Conference on Learning Representations},
  year={2023},
}

@inproceedings{radford2021learning,
  title={Learning transferable visual models from natural language supervision},
  author={Radford, A. and Kim, J. W. and Hallacy, C. and Ramesh, A. and Goh, G. and Agarwal, S. and Sutskever, I.},
  booktitle={Proceedings of the International Conference on Machine Learning},
  pages={8748--8763},
  year={2021},
}

@inproceedings{wang2024real,
  title={Real-iad: A real-world multi-view dataset for benchmarking versatile industrial anomaly detection},
  author={Wang, Chengjie and Zhu, Wenbing and Gao, Bin-Bin and Gan, Zhenye and Zhang, Jiangning and Gu, Zhihao and Qian, Shuguang and Chen, Mingang and Ma, Lizhuang},
  booktitle={Proceedings of the IEEE/CVF Conference on Computer Vision and Pattern Recognition},
  pages={22883--22892},
  year={2024}
}

@article{pang2025context,
  title={Context-aware feature reconstruction for class-incremental anomaly detection and localization},
  author={Pang, Jingxuan and Li, Chunguang},
  journal={Neural Networks},
  volume={181},
  pages={106788},
  year={2025},
  publisher={Elsevier}
}

@inproceedings{bang2021rainbow,
  title={Rainbow memory: Continual learning with a memory of diverse samples},
  author={Bang, Jihwan and Kim, Heesu and Yoo, YoungJoon and Ha, Jung-Woo and Choi, Jonghyun},
  booktitle={Proceedings of the IEEE/CVF conference on computer vision and pattern recognition},
  pages={8218--8227},
  year={2021}
}

@inproceedings{deng2009imagenet,
  title={Imagenet: A large-scale hierarchical image database},
  author={Deng, Jia and Dong, Wei and Socher, Richard and Li, Li-Jia and Li, Kai and Fei-Fei, Li},
  booktitle={2009 IEEE conference on computer vision and pattern recognition},
  pages={248--255},
  year={2009},
  organization={Ieee}
}

@inproceedings{lehr2024ad3,
  title={AD3: Introducing a Score for Anomaly Detection Dataset Difficulty Assessment Using VIADUCT Dataset},
  author={Lehr, Jan and Philipps, Jan and Sargsyan, Alik and Pape, Martin and Kr{\"u}ger, J{\"o}rg},
  booktitle={European Conference on Computer Vision},
  pages={449--464},
  year={2024},
  organization={Springer}
}

@inproceedings{zhu2024toward,
  title={Toward generalist anomaly detection via in-context residual learning with few-shot sample prompts},
  author={Zhu, Jiawen and Pang, Guansong},
  booktitle={Proceedings of the IEEE/CVF Conference on Computer Vision and Pattern Recognition},
  pages={17826--17836},
  year={2024}
}

@inproceedings{mcintosh2024unsupervised,
  title={Unsupervised, Online and On-The-Fly Anomaly Detection for Non-stationary Image Distributions},
  author={McIntosh, Declan and Albu, Alexandra Branzan},
  booktitle={European Conference on Computer Vision},
  pages={428--445},
  year={2024},
  organization={Springer}
}

@inproceedings{meng2024moead,
  title={MoEAD: A Parameter-Efficient Model for Multi-class Anomaly Detection},
  author={Meng, Shiyuan and Meng, Wenchao and Zhou, Qihang and Li, Shizhong and Hou, Weiye and He, Shibo},
  booktitle={European Conference on Computer Vision},
  pages={345--361},
  year={2024},
  organization={Springer}
}

@inproceedings{qu2024vcp,
  title={VCP-CLIP: A visual context prompting model for zero-shot anomaly segmentation},
  author={Qu, Zhen and Tao, Xian and Prasad, Mukesh and Shen, Fei and Zhang, Zhengtao and Gong, Xinyi and Ding, Guiguang},
  booktitle={European Conference on Computer Vision},
  pages={301--317},
  year={2024},
  organization={Springer}
}

@inproceedings{gui2024few,
  title={Few-shot anomaly-driven generation for anomaly classification and segmentation},
  author={Gui, Guan and Gao, Bin-Bin and Liu, Jun and Wang, Chengjie and Wu, Yunsheng},
  booktitle={European Conference on Computer Vision},
  pages={210--226},
  year={2024},
  organization={Springer}
}

@article{tao2024kernel,
  title={Kernel-Aware Graph Prompt Learning for Few-Shot Anomaly Detection},
  author={Tao, Fenfang and Xie, Guo-Sen and Zhao, Fang and Shu, Xiangbo},
  journal={arXiv preprint arXiv:2412.17619},
  year={2024}
}

@inproceedings{zhang2024mediclip,
  title={Mediclip: Adapting clip for few-shot medical image anomaly detection},
  author={Zhang, Ximiao and Xu, Min and Qiu, Dehui and Yan, Ruixin and Lang, Ning and Zhou, Xiuzhuang},
  booktitle={International Conference on Medical Image Computing and Computer-Assisted Intervention},
  pages={458--468},
  year={2024},
  organization={Springer}
}

@inproceedings{chen2024unified,
  title={A unified anomaly synthesis strategy with gradient ascent for industrial anomaly detection and localization},
  author={Chen, Qiyu and Luo, Huiyuan and Lv, Chengkan and Zhang, Zhengtao},
  booktitle={European Conference on Computer Vision},
  pages={37--54},
  year={2024},
  organization={Springer}
}

@inproceedings{mishra2021vt,
  title={VT-ADL: A vision transformer network for image anomaly detection and localization},
  author={Mishra, Pankaj and Verk, Riccardo and Fornasier, Daniele and Piciarelli, Claudio and Foresti, Gian Luca},
  booktitle={2021 IEEE 30th International Symposium on Industrial Electronics (ISIE)},
  pages={01--06},
  year={2021},
  organization={IEEE}
}

@inproceedings{he2024learning,
  title={Learning Unified Reference Representation for Unsupervised Multi-class Anomaly Detection},
  author={He, Liren and Jiang, Zhengkai and Peng, Jinlong and Zhu, Wenbing and Liu, Liang and Du, Qiangang and Hu, Xiaobin and Chi, Mingmin and Wang, Yabiao and Wang, Chengjie},
  booktitle={European Conference on Computer Vision},
  pages={216--232},
  year={2024},
  organization={Springer}
}

@inproceedings{gu2024filo,
  title={Filo: Zero-shot anomaly detection by fine-grained description and high-quality localization},
  author={Gu, Zhaopeng and Zhu, Bingke and Zhu, Guibo and Chen, Yingying and Li, Hao and Tang, Ming and Wang, Jinqiao},
  booktitle={Proceedings of the 32nd ACM International Conference on Multimedia},
  pages={2041--2049},
  year={2024}
}

@article{tamura2023random,
  title={Random word data augmentation with clip for zero-shot anomaly detection},
  author={Tamura, Masato},
  journal={arXiv preprint arXiv:2308.11119},
  year={2023}
}

@article{chen2023zero,
  title={A zero-/fewshot anomaly classification and segmentation method for cvpr 2023 vand workshop challenge tracks 1\&2: 1st place on zero-shot ad and 4th place on few-shot ad},
  author={Chen, Xuhai and Han, Yue and Zhang, Jiangning},
  journal={arXiv preprint arXiv:2305.17382},
  volume={2},
  number={4},
  year={2023}
}

@inproceedings{chen2024clip,
  title={Clip-ad: A language-guided staged dual-path model for zero-shot anomaly detection},
  author={Chen, Xuhai and Zhang, Jiangning and Tian, Guanzhong and He, Haoyang and Zhang, Wuhao and Wang, Yabiao and Wang, Chengjie and Liu, Yong},
  booktitle={International Joint Conference on Artificial Intelligence},
  pages={17--33},
  year={2024},
  organization={Springer}
}

@article{deng2023anovl,
  title={Anovl: Adapting vision-language models for unified zero-shot anomaly localization},
  author={Deng, Hanqiu and Zhang, Zhaoxiang and Bao, Jinan and Li, Xingyu},
  journal={arXiv preprint arXiv:2308.15939},
  year={2023}
}

@inproceedings{zavrtanik2021draem,
  title={Draem-a discriminatively trained reconstruction embedding for surface anomaly detection},
  author={Zavrtanik, Vitjan and Kristan, Matej and Sko{\v{c}}aj, Danijel},
  booktitle={Proceedings of the IEEE/CVF international conference on computer vision},
  pages={8330--8339},
  year={2021}
}

@inproceedings{li2024promptad,
  title={Promptad: Learning prompts with only normal samples for few-shot anomaly detection},
  author={Li, Xiaofan and Zhang, Zhizhong and Tan, Xin and Chen, Chengwei and Qu, Yanyun and Xie, Yuan and Ma, Lizhuang},
  booktitle={Proceedings of the IEEE/CVF Conference on Computer Vision and Pattern Recognition},
  pages={16838--16848},
  year={2024}
}

@inproceedings{ristea2022self,
  title={Self-supervised predictive convolutional attentive block for anomaly detection},
  author={Ristea, Nicolae-C{\u{a}}t{\u{a}}lin and Madan, Neelu and Ionescu, Radu Tudor and Nasrollahi, Kamal and Khan, Fahad Shahbaz and Moeslund, Thomas B and Shah, Mubarak},
  booktitle={Proceedings of the IEEE/CVF conference on computer vision and pattern recognition},
  pages={13576--13586},
  year={2022}
}

@inproceedings{li2021cutpaste,
  title={Cutpaste: Self-supervised learning for anomaly detection and localization},
  author={Li, Chun-Liang and Sohn, Kihyuk and Yoon, Jinsung and Pfister, Tomas},
  booktitle={Proceedings of the IEEE/CVF conference on computer vision and pattern recognition},
  pages={9664--9674},
  year={2021}
}

@article{cohen2020sub,
  title={Sub-image anomaly detection with deep pyramid correspondences},
  author={Cohen, Niv and Hoshen, Yedid},
  journal={arXiv preprint arXiv:2005.02357},
  year={2020}
}

@article{bergmann2022beyond,
  title={Beyond dents and scratches: Logical constraints in unsupervised anomaly detection and localization},
  author={Bergmann, Paul and Batzner, Kilian and Fauser, Michael and Sattlegger, David and Steger, Carsten},
  journal={International Journal of Computer Vision},
  volume={130},
  number={4},
  pages={947--969},
  year={2022},
  publisher={Springer}
}

@article{jin2024oner,
  title={ONER: Online Experience Replay for Incremental Anomaly Detection},
  author={Jin, Yizhou and Zhu, Jiahui and Wang, Guodong and Li, Shiwei and Zhang, Jinjin and Liu, Qingjie and Liu, Xinyue and Wang, Yunhong},
  journal={arXiv preprint arXiv:2412.03907},
  year={2024}
}

@inproceedings{tang2024incremental,
  title={An incremental unified framework for small defect inspection},
  author={Tang, Jiaqi and Lu, Hao and Xu, Xiaogang and Wu, Ruizheng and Hu, Sixing and Zhang, Tong and Cheng, Tsz Wa and Ge, Ming and Chen, Ying-Cong and Tsung, Fugee},
  booktitle={European conference on computer vision},
  pages={307--324},
  year={2024},
  organization={Springer}
}

@inproceedings{li2022towards,
  title={Towards continual adaptation in industrial anomaly detection},
  author={Li, Wujin and Zhan, Jiawei and Wang, Jinbao and Xia, Bizhong and Gao, Bin-Bin and Liu, Jun and Wang, Chengjie and Zheng, Feng},
  booktitle={Proceedings of the 30th ACM International Conference on Multimedia},
  pages={2871--2880},
  year={2022}
}

@inproceedings{liu2024unsupervised,
  title={Unsupervised Continual Anomaly Detection with Contrastively-Learned Prompt},
  author={Liu, Jiaqi and Wu, Kai and Nie, Qiang and Chen, Ying and Gao, Bin-Bin and Liu, Yong and Wang, Jinbao and Wang, Chengjie and Zheng, Feng},
  booktitle={Proceedings of the AAAI Conference on Artificial Intelligence},
  volume={38},
  pages={3639--3647},
  year={2024}
}

@inproceedings{yao2024resad,
  title     = {ResAD: A Simple Framework for Class Generalizable Anomaly Detection},
  author    = {Yao, Xincheng and Chen, Ziqi and Gao, Cheng and Zhai, Guangtao and Zhang, Caiming},
  booktitle = {Advances in Neural Information Processing Systems},
  volume    = {37},
  pages     = {125287--125311},
  year      = {2024}
}

@article{wei2025deep,
  title={Deep learning-based anomaly detection for precision field crop protection},
  author={Wei, Cheng and Shan, Yifeng and Zhen, MengZhe},
  journal={Frontiers in Plant Science},
  volume={16},
  pages={1576756},
  year={2025},
  publisher={Frontiers Media SA}
}

@article{guo2025two,
  title={A Two-Stage Deep-Learning Framework for Industrial Anomaly Detection: Integrating Small-Sample Semantic Segmentation and Knowledge Distillation},
  author={Guo, Lei and Lv, Feiya},
  journal={Machines},
  volume={13},
  number={8},
  pages={712},
  year={2025},
  publisher={Multidisciplinary Digital Publishing Institute}
}

@inproceedings{fang2023fastrecon,
  title={Fastrecon: Few-shot industrial anomaly detection via fast feature reconstruction},
  author={Fang, Zheng and Wang, Xiaoyang and Li, Haocheng and Liu, Jiejie and Hu, Qiugui and Xiao, Jimin},
  booktitle={Proceedings of the IEEE/CVF International Conference on Computer Vision},
  pages={17481--17490},
  year={2023}
}

@article{wang2024comprehensive,
  title={A comprehensive survey of continual learning: Theory, method and application},
  author={Wang, Liyuan and Zhang, Xingxing and Su, Hang and Zhu, Jun},
  journal={IEEE transactions on pattern analysis and machine intelligence},
  volume={46},
  number={8},
  pages={5362--5383},
  year={2024},
  publisher={IEEE}
}

@inproceedings{bugarin2024unveiling,
  title={Unveiling the anomalies in an ever-changing world: A benchmark for pixel-level anomaly detection in continual learning},
  author={Bugarin, Nikola and Bugaric, Jovana and Barusco, Manuel and Pezze, Davide Dalle and Susto, Gian Antonio},
  booktitle={Proceedings of the IEEE/CVF Conference on Computer Vision and Pattern Recognition},
  pages={4065--4074},
  year={2024}
}

@inproceedings{kingma2014adam,
  title   = {Adam: A Method for Stochastic Optimization},
  author  = {Kingma, Diederik P. and Ba, Jimmy},
  booktitle = {International Conference on Learning Representations},
  year    = {2015}
}
\bibliographystyle{icml2026}

\newpage
\appendix
\onecolumn
\clearpage
\setcounter{table}{0}
\renewcommand{\thetable}{\Alph{table}}
\setcounter{figure}{0}
\renewcommand{\thefigure}{\Alph{figure}}


\section{Continual-MEGA Benchmark Details}

\subsection{ContinualAD Dataset}
\begin{figure}[h]
\centering
\includegraphics[width=\textwidth]
{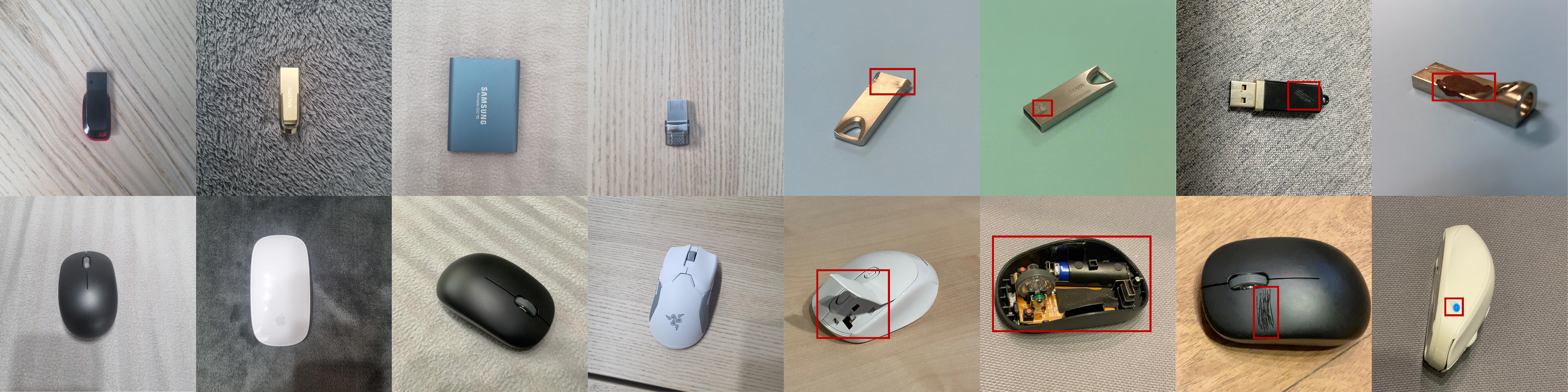}
\caption{\textbf{Example visualization of ContinualAD dataset.} For comprehensive benchmarking across diverse environments, the ContinualAD dataset was curated to encompass images featuring a wide range of backgrounds. The red boxes indicate the anomaly regions.}
\label{fig:dataset_visualization}
\end{figure}

Figure~\ref{fig:dataset_visualization} illustrates sample images from the ContinualAD dataset, which includes diverse scenes captured in different background settings.
Table~\ref{tab:cls_counts} presents the class names along with the numbers of normal and anomaly samples for each class.
The red boxes highlight the anomalous regions. Moreover, the ContinualAD dataset provides multiple instances and varied backgrounds even within the same object class, enabling a more comprehensive evaluation of anomaly detection performance compared to previously released datasets.

\begin{table}[h]
\centering
\resizebox{0.65\linewidth}{!}{
\begin{tabular}{l c c c l c c}
\Xhline{2\arrayrulewidth}
Class & \#Normal & \#Anomaly & & Class & \#Normal & \#Anomaly \\
\hline
Energy-bar & 329 & 542 & & Toy & 368 & 492 \\
Apple & 490 & 502 & & Multi-pen & 494 & 492 \\
Kleenex & 480 & 519 & & Chopsticks & 488 & 524 \\
Ruler & 277 & 490 & & Watermelon & 497 & 506 \\
Toothpaste & 513 & 516 & & Egg & 662 & 491 \\
Sunglasses & 499 & 572 & & Spoon & 527 & 517 \\
Capsule & 507 & 493 & & Calculator & 506 & 500 \\
Flash-drive & 522 & 495 & & Eraser & 458 & 508 \\
Band-aid & 511 & 491 & & Mango & 456 & 491 \\
Cucumber & 505 & 507 & & Candy & 517 & 490 \\
Toothbrush & 500 & 549 & & Mouse & 517 & 495 \\
Soap & 394 & 787 & & Glasses-case & 501 & 549 \\
Dollar & 391 & 494 & & Notebook & 354 & 513 \\
Pencil & 518 & 517 & & Food-container & 520 & 489 \\
Fork & 537 & 507 & & Cup & 517 & 488 \\
\Xhline{2\arrayrulewidth}
\end{tabular}
}
\caption{\textbf{Number of normal and anomaly samples} per class in the ContinualAD dataset.}
\label{tab:cls_counts}
\end{table}

\subsection{Scenario Overview and Class Distribution}
Table~\ref{tab:scenario_overview} provides an overview of the three evaluation scenarios in the Continual-MEGA benchmark.
Scenario~1 uses all seven datasets in the continual stream.
Among them, 85 classes are assigned to the base task, and the remaining 60 classes are introduced as new classes.
We consider task sizes of 5, 10, and 30 classes (85–5/10/30).
Scenario~2 is designed to evaluate zero-shot performance after continual adaptation.
To create a held-out target, MVTec-AD and VisA are removed from the training stream.
In this case, the base task consists of 58 classes, and 60 additional classes are used as continual new classes.
Zero-shot evaluation is then performed on MVTec-AD and VisA.
Scenario~3 examines how the newly collected ContinualAD dataset affects zero-shot performance.
Here, ContinualAD is excluded from the adaptation stream, while the other datasets are kept as in Scenario~2.
The base task again contains 58 classes, but only 30 classes are used as continual new classes.
As in Scenario~2, zero-shot performance is evaluated on MVTec-AD and VisA under this reduced-diversity stream.

\begin{figure*}[t]
\centering
    \centering
    \includegraphics[width=\textwidth]{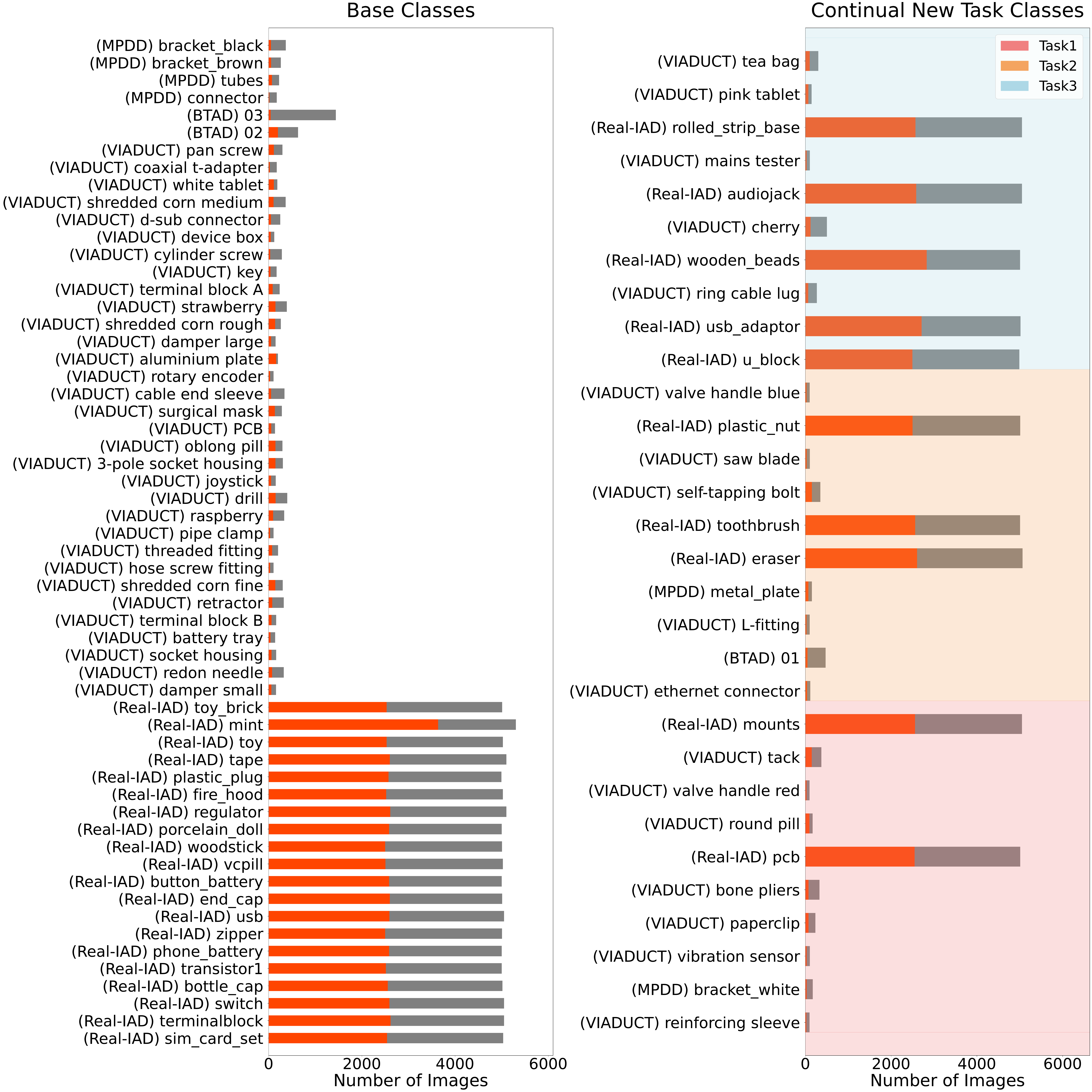}
    \caption{Class distribution of Scenario 3. This example illustrates Scenario 3 when each task contains 10 classes, and three tasks arrive sequentially.
The seven colored background bands indicate one \textit{Base} block (left) and the six incremental \textit{New} task blocks (right) that arrive in order.
The orange line denotes the anomaly-sample count per class, and the gray bars denote the total sample volume.}
    \label{fig:scenario3_full}
\end{figure*}

Figures~\ref{fig:scenario1_full},~\ref{fig:scenario2_full}, and~\ref{fig:scenario3_full} illustrate the detailed class distributions for Scenario 1, 2 and 3 of the Continual-MEGA Benchmark, respectively. We observe a notable imbalance among classes in all scenarios.
This imbalance underscores the challenge of maintaining consistent anomaly detection performance, as smaller classes from previous datasets are more prone to being forgotten during adaptation. In particular, Scenario 3 excludes the ContinualAD dataset, leading to fewer samples and a more constrained class distribution compared to Scenario 2. Such variations in class composition and sample availability across the scenarios make them valuable for comprehensively evaluating the robustness and adaptability of continual anomaly detection methods.

\begin{table}[h]
\centering
\resizebox{\linewidth}{!}{
\begin{tabular}{lccc}
\Xhline{2\arrayrulewidth}
\multirow{2}{*}{Scenario (Base-New)} & Base classes  & New classes stream  & \multirow{2}{*}{Zero-shot evaluation} \\
& (datasets) & (datasets / \#classes) & \\
\hline
Scenario 1 (85-5/10/30) 
& All 7 Datasets
& Same as base / 60 classes 
& -- \\
Scenario 2 (58-5/10/30) 
& excluding MVTec-AD, VisA
& Same as base / 60 classes 
& MVTec-AD, VisA \\
Scenario 3 (58-5/10/30) 
& excluding MVTec-AD, VisA, ContinualAD
& Same as base / 30 classes 
& MVTec-AD, VisA \\
\Xhline{2\arrayrulewidth}
\end{tabular}
}
\caption{Overview of the three evaluation scenarios in the Continual-MEGA benchmark. 
For each scenario, we summarize the datasets used for the base classes, the continual
(New) class stream, and any datasets that are held out for zero-shot evaluation.}
\label{tab:scenario_overview}
\end{table}

\begin{table}[h]
\centering
\resizebox{0.7\textwidth}{!}{
\begin{tabular}{lcccccc}
\Xhline{2\arrayrulewidth}
\multirow{2}{*}{\textbf{Scenario}} & \multirow{2}{*}{\textbf{Stage}} & \multicolumn{2}{c}{\textbf{Train}} & \multicolumn{2}{c}{\textbf{Test}} \\
& & \#Normal & \#Anomaly & \#Normal & \#Anomaly \\
\hline
\multirow{2}{*}{Scenario 1} 
& Base               & 850  & 850  & 71,274 & 44,301 \\
& Continual & 600  & 600  & 49,543 & 28,140 \\
\hline
\multirow{2}{*}{Scenario 2} 
& Base               & 580  & 580  & 59,121 & 35,972 \\
& Continual & 600  & 600  & 60,267 & 34,281 \\
\hline
\multirow{2}{*}{Scenario 3} 
& Base               & 580  & 580  & 69,788 & 37,130 \\
& Continual  & 300  & 300  & 35,245 & 17,597 \\
\Xhline{2\arrayrulewidth}
\end{tabular}
}
\caption{\textbf{Number of training and test samples in each scenario.}}
\label{tab:scenario_sample_counts}
\end{table}

\begin{figure*}[h]
\centering
\includegraphics[width=0.7\linewidth]{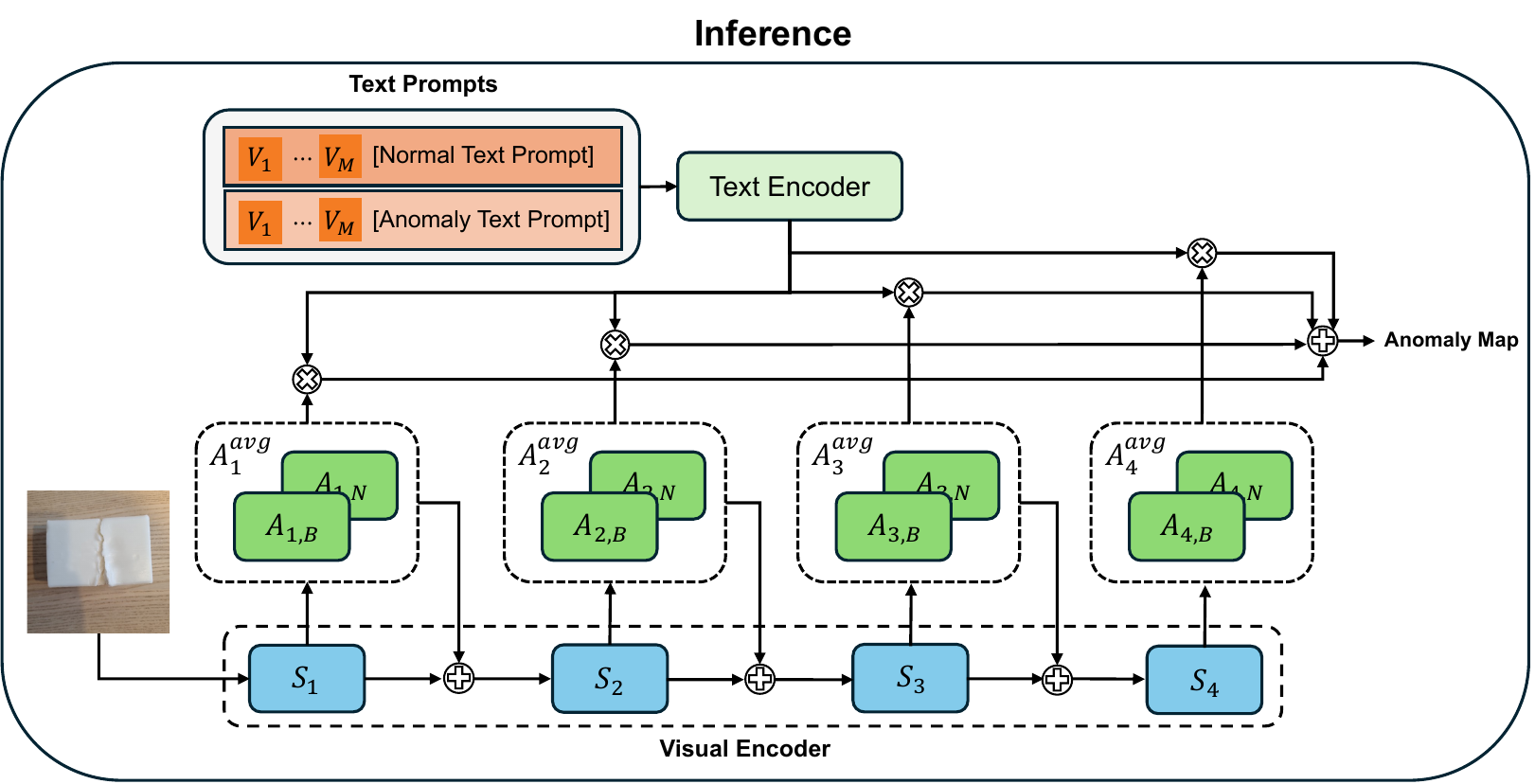}
\caption{\textbf{Overview architecture of inference process.} $B$ and $N$ denote the adapters corresponding to the base classes and the task-specific classes $\{1, 2, 3, ..., N\}$, respectively. During inference, we use $A^{\text{avg}}$, which is the average of the adapter weights trained on the base classes and the $N$ task adapters.}
\label{fig:overview_inference}
\end{figure*}

\subsection{Training Setup for Continual-MEGA Benchmark}
To simulate a low-resource environment where training samples are limited, we adopt a minimal supervision setting in the proposed Continual-MEGA benchmark. Specifically, only 10 normal and 10 anomalous training images are provided per class during both the base and continual learning stages.
Table~\ref{tab:scenario_sample_counts} shows the number of training and test images used for base classes and continual learning classes in each scenario.
For continual learning, the classes are partitioned into three task settings, with each task comprising 5, 15, and 30 classes, respectively, to simulate varying levels of incremental difficulty.
\textit{Following recent trends in AD toward few-/zero-shot and continual learning, we adopt a limited number of training and adaptation samples to better reflect realistic constraints.}
However, depending on the target AD application, the training setup of our Continual-MEGA benchmark can be flexibly reconfigured to suit different deployment scenarios. This will be discussed in the Limitations section in more detail.

Figure~\ref{fig:overview_training} and Figure~\ref{fig:overview_inference} illustrate an overview of the proposed baseline method, depicting the training and inference processes.
During training, the model incorporates Mixture-of-Experts adapters to dynamically specialize representations across tasks, while synthetic anomaly feature generation is employed to enrich limited samples and improve adaptation.
Additionally, we leverage generalized text prompts to obtain text features that are not specific to any particular domain or class.
Table~\ref{tab:text_prompts} lists the prompts used to extract these generalized features, consisting of 10 prompts each for normal and anomaly classes.
This design helps the model capture more robust, domain-agnostic representations that improve anomaly detection performance across continual learning tasks.

For the proposed baseline method, hyperparameter tuning was conducted solely on the base classes of Scenario 1 in a lightweight manner. The resulting hyperparameters were uniformly used across all remaining scenarios to ensure fair and consistent evaluation.
To evaluate model performance in a realistic continual learning setting, we avoided scenario-specific hyperparameter tuning. This design choice aims to reflect practical constraints in real-world deployments, where careful tuning for each newly incoming task is often infeasible~\citep{cha2024hyperparameters}.

\begin{table*}[t]
\centering
\resizebox{0.8\textwidth}{!}
{
\begin{tabular}{c|c|c}
\Xhline{2\arrayrulewidth}
Prompt No. &Normal Prompts & Anomaly Prompts \\ \hline
1&This is an example of a normal object & This is an example of an anomalous object \\
2&This is a typical appearance of the object & This is not the typical appearance of the object \\
3&This is what a normal object looks like & This is what an anomaly looks like \\
4&A photo of a normal object & A photo of an anomalous object \\
5&This is not an anomaly & This is an example of an abnormal object \\
6&This is an example of a standard object & This is an example of an abnormal object \\
7&This is the standard appearance of the object & This is not the usual appearance of the object \\
8&This is what a standard object looks like & This is what an abnormal object looks like \\
9&A photo of a standard object & A photo of an abnormal object \\
10&This object meets standard characteristics & An abnormality detected in this object \\

\Xhline{2\arrayrulewidth}
\end{tabular}
}
\caption{\textbf{Description of generalized text prompts.}}
\label{tab:text_prompts}
\end{table*}

\begin{table*}[!t]
\centering
\resizebox{0.8\textwidth}{!}{
\begin{tabular}{c|c|cc|cc|cc}
\Xhline{2\arrayrulewidth}
\multirow{2}{*}{Type}& \multirow{2}{*}{Method} & \multicolumn{2}{c|}{Scenario 1 (85 classes)} & \multicolumn{2}{c|}{Scenario 2 (58 classes)} & \multicolumn{2}{c}{Scenario 3 (58 classes)} \\ 
&& \makebox[3.8em][c]{Image} & Pixel & \makebox[3.8em][c]{Image} & Pixel & \makebox[3.8em][c]{Image} & Pixel \\ \hline
\multirow{4}{*}{Only-normal} &SimpleNet      & 58.8 & 6.3 & 61.3 & 4.5 & 57.5 & 4.5 \\
&GeneralAD     & 51.5 & 2.6 & 52.6 & 1.8 & 54.4 & 2.7 \\
&HGAD          & 59.5 & 5.0 & 56.1 & 3.2 & 55.5 & 2.7 \\
&ResAD    & 73.3 & 15.5 & 69.1 & 7.8 & 70.7 & 15.3 \\ \hline
\multirow{3}{*}{VLM-based} &MVFA          & 81.7 & 32.6 & 65.8 & 10.4 & 70.7 & 21.2 \\
&VCP-CLIP      & 73.8 & 25.4 & 61.0 & 23.1 & 61.9 & 22.5 \\
&MediCLIP      & 73.9 & 4.5 & 78.1 & 8.5 & 75.3 & 5.9 \\\hline
\multirow{4}{*}{Continual} &UCAD         & 55.8 & 1.6 & 58.1 & 4.7 & 56.0 & 3.6 \\
&IUF          & 60.5 & 7.4 & 57.4 & 4.4 & 58.5 & 4.2 \\ 
&$\text{IUF}^\ast$    & 68.3 & 13.5 & 65.8 & 9.5 & 63.6 & 9.5 \\
&\textbf{Ours} & \textbf{83.1} & \textbf{39.0} & \textbf{82.0} & \textbf{35.7} & \textbf{77.8} & \textbf{36.5} \\

\Xhline{2\arrayrulewidth}
\end{tabular}
}

\caption{\textbf{Experimental results of base classes across scenarios.}
We note that the base classes used in Scenario 2 and Scenario 3 differ, as ContinualAD is included among the base classes in Scenario 2 but not in Scenario 3. The best-performing results are highlighted in \textbf{bold}.}
\label{tab:base classes}
\end{table*}

\begin{figure*}[t]
\centering
\includegraphics[width=\textwidth]{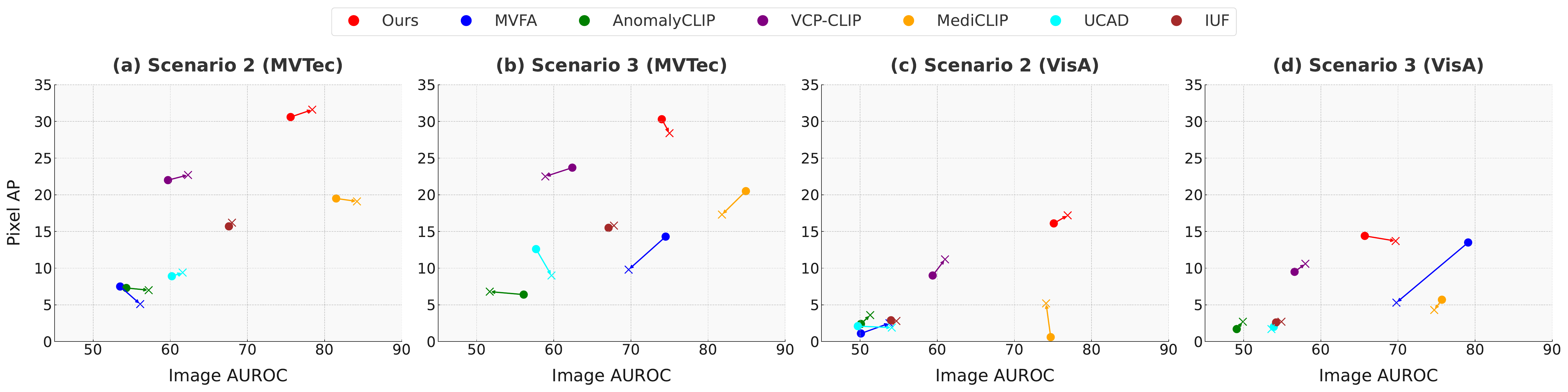}
\caption{\textbf{Image-level AUROC and pixel-level AP performance on MVTec-AD (a, b) and VisA (c, d) datasets.} Each point represents the performance of a method before (\textbullet) and after ($\times$) continual learning. Arrows indicate the performance change from the model trained only on \textit{Base} classes to the model trained via continual learning. The continual learning results are averaged over three settings, where each task consists of 5, 10, and 30 \textit{New} classes, respectively.
Notably, in Scenario 3, where the proposed the ContinualAD dataset is excluded, most methods experience a noticeable drop in continual zero-shot learning (CZSL) performance, 
highlighting the importance of incorporating ContinualAD for robust generalization.
}
\label{fig:base_continual_zero}
\vspace{-1.0em}
\end{figure*}


\subsection{Continual-MEGA Benchmark Evaluation Details}
\label{app:eval}
This section provides an in-depth analysis of the evaluation results of various AD methods on our Continual-MEGA Benchmark, complementing the main experiments.
We note that Scenario 3 excludes the ContinualAD dataset from both the \textit{Base} and \textit{New} classes of the benchmark.
Figure~\ref{fig:base_continual_zero} of the paper compares the zero-shot performance of various methods on the MVTec-AD and VisA datasets under two conditions: (1) trained only on the \textit{Base} classes, and (2) after continual adaptation as defined by our proposed Continual-MEGA benchmark.

From the results presented, we observe that anomaly detection (AD) performance has the following tendencies: 
(1) Compared to Scenario 3, overall AD performance improves in Scenario 2 across most methods, showing the effectiveness of the ContinualAD dataset.
(2) Continual adaptation using the ContinualAD dataset enhances zero-shot generalizability, as observed in MVTec-AD and VisA. Excluding ContinualAD leads to a consistent drop in performance among prior methods.
(3) Our proposed baseline achieves strong and consistent results across all scenarios, showing notable improvements in Scenario 2 and maintaining competitive generalizability in Scenario 3 after continual adaptation.


Additionally, performing anomaly detection on the baseline categories in our proposed setup is substantially more challenging than in conventional benchmarks, as shown in Table~\ref{tab:base classes}. Under this more difficult setting, VLM-based methods demonstrate significantly stronger performance compared to approaches explicitly designed for continual learning. This performance gap stems from the inherently limited detection capability of existing continual anomaly detection methods at their initial stage.

\section{Implementation Details}
We use CLIP with ViT-L/14~\citep{dosovitskiy2020image} architecture, which consists of 24 sublayers divided into four layers, where each layer contains six sublayers. The size of input images was set to 336. The adaptation layers for anomaly feature generation were applied to layers 1, 2, 3, and 4. The batch size was set to 16. The adapter parameters are optimized using AdamW~\citep{loshchilov2017decoupled} with a learning rate of $1\times10^{-4}$ for 50 epochs on the $Base$ classes and for 20 epochs for each continual task under the same optimization setup. The learnable text-prompt module is trained only on the $Base$ classes using Adam~\citep{kingma2014adam} with a learning rate of $1\times10^{-4}$ and is kept frozen during all continual tasks. For synthetic anomaly feature generation, we use the random noise term $\gamma$ as Gaussian noise with standard deviation $0.25$, i.e., $\gamma \sim \mathcal{N}(0, 0.25^2)$, and add it to the feature space.

\section{Discussions: Limitations and Future Work}
\textbf{Regarding the dataset sample configuration}, a primary limitation of the proposed benchmark is class imbalance, as sample sizes vary significantly across datasets. In our continual evaluation setup, models that better fit classes having a smaller number of samples would be beneficial to achieve higher performance.
While this setting reflects the class imbalance observed in real-world inspection, balancing sample quantities across classes would improve the reliability of AD performance evaluation in the continual setup.

\textbf{Regarding the benchmark training and evaluation configuration}, the Continual-MEGA benchmark intentionally adopts limited training and adaptation samples to evaluate the effectiveness of recent few-/zero-shot AD methods under both unseen and continual setups, leveraging a significantly larger evaluation set. 
While our primary focus is evaluation, we expect higher accuracy with increased training data, particularly for the \textit{Base} set, making detailed analysis across varying training sizes a key future direction.

\textbf{From a model perspective}, our baseline, despite its simplicity, achieves strong performance across diverse scenarios in the Continual-MEGA benchmark. As this work focuses primarily on benchmark construction, deeper analysis through ablations and developing improved AD methods remain essential directions for future research.

\begin{figure*}[t]
\centering
    \centering
    \includegraphics[width=0.8\textwidth]{Figure/class_distribution_scenario1_horizontal2_75.pdf}
    \caption{Class distribution of Scenario 1. This example illustrates Scenario 1 when each task contains 10 classes, and six tasks arrive sequentially.
The seven colored background bands indicate one \textit{Base} block (left) and the six incremental \textit{New} task blocks (right) that arrive in order.
The orange line denotes the anomaly-sample count per class, and the gray bars denote the total sample volume.}
    \label{fig:scenario1_full}
\end{figure*}

\begin{figure*}[t]
\centering
    \centering
    \includegraphics[width=0.8\textwidth]{Figure/class_distribution_scenario2_horizontal2_75.pdf}
    \caption{Class distribution of Scenario 2. This example illustrates Scenario 2 when each task contains 10 classes, and six tasks arrive sequentially.
The seven colored background bands indicate one \textit{Base} block (left) and the six incremental \textit{New} task blocks (right) that arrive in order.
The orange line denotes the anomaly-sample count per class, and the gray bars denote the total sample volume.}
    \label{fig:scenario2_full}
\end{figure*}

\clearpage

\end{document}